\documentclass[letterpaper, 10 pt, journal, twoside]{IEEEtran}  % Use this command for final RAL version

\IEEEoverridecommandlockouts                              % This command is only needed if
\usepackage{units}
\usepackage{array}
\usepackage{etex} 
\usepackage{multicol}
\usepackage{hyperref}

\usepackage{float}

\usepackage{graphicx}
\usepackage{epstopdf}
\usepackage{epsfig}
\usepackage{xspace}

\usepackage{enumitem}
\usepackage{amsmath,bm}
\usepackage{amsmath,amssymb}
\usepackage{graphicx}
\usepackage[colorinlistoftodos]{todonotes}
\newcommand{\journalVersion}[1]{}
\usepackage{xr}
\graphicspath{{./figures/}}
\usepackage{comment}

\newtheorem{theorem}{Theorem}

\newtheorem{definition}[theorem]{Definition}

\newtheorem{remark}[theorem]{Remark}

\newcommand{\bdmath}{\begin{dmath}}
\newcommand{\edmath}{\end{dmath}}
\newcommand{\beq}{\begin{equation}}
\newcommand{\eeq}{\end{equation}}
\newcommand{\bdm}{\begin{displaymath}}
\newcommand{\edm}{\end{displaymath}}
\newcommand{\bea}{\begin{eqnarray}}
\newcommand{\eea}{\end{eqnarray}}
\newcommand{\beal}{\beq \begin{array}{lll}}
\newcommand{\eeal}{\end{array} \eeq}
\newcommand{\beas}{\begin{eqnarray*}}
\newcommand{\eeas}{\end{eqnarray*}}
\newcommand{\ba}{\begin{array}}
\newcommand{\ea}{\end{array}}
\newcommand{\bit}{\begin{itemize}}
\newcommand{\eit}{\end{itemize}}
\newcommand{\ben}{\begin{enumerate}}
\newcommand{\een}{\end{enumerate}}

\newcommand{\calC}{{\cal C}}
\newcommand{\calD}{{\cal D}}

\newcommand{\calM}{{\cal M}}
\newcommand{\calN}{{\cal N}}

\newcommand{\calS}{{\cal S}}

\newcommand{\calX}{{\cal X}}

\newcommand{\M}[1]{{\bm #1}} % Face for matrices
\renewcommand{\boldsymbol}[1]{{\bm #1}}
\newcommand{\hide}[1]{}

\newcommand{\hiddenText}{{\color{gray} hidden text.}}
\newcommand{\hideWithText}[1]{\hiddenText}

\newcommand{\subject}{\text{ subject to }}

\newcommand{\tran}{^{\mathsf{T}}}

\newcommand{\e}{{\mathrm e}}

\newcommand{\ones}{{\mathbf 1}}

\newcommand{\Real}[1]{ { {\mathbb R}^{#1} } }

\newcommand{\at}[1]{^{(#1)}}

\newcommand{\setdef}[2]{ \{#1 \; {:} \; #2 \} }

\newcommand{\MM}{\M{M}}

\newcommand{\MF}{\M{F}}

\newcommand{\MZ}{\M{Z}}

\newcommand{\vv}{\boldsymbol{v}}

\newcommand{\vxx}{\boldsymbol{x}} 
\newcommand{\vy}{\boldsymbol{y}}

\newcommand{\blue}[1]{{\color{blue}#1}}

\newcommand{\linkToPdf}[1]{\href{#1}{\blue{(pdf)}}}
\newcommand{\linkToPpt}[1]{\href{#1}{\blue{(ppt)}}}
\newcommand{\linkToCode}[1]{\href{#1}{\blue{(code)}}}
\newcommand{\linkToWeb}[1]{\href{#1}{\blue{(web)}}}
\newcommand{\linkToVideo}[1]{\href{#1}{\blue{(video)}}}
\newcommand{\award}[1]{\xspace} % {{\red{#1}}} % omit awards

\newcommand{\vz}{\boldsymbol{z}}

\renewcommand{\MM}{\mathcal{M}}

\newcommand{\myparagraph}[1]{{\bf #1.}}

\newcommand{\myParagraph}[1]{{\bf #1.}}
\newcommand{\mvec}{m} % module vectors
\newcommand{\mvecb}{\hat{m}} % module vectors
\newcommand{\fvec}{f} % module vectors
\newcommand{\dvar}{a} % discrete variables
\newcommand{\bvar}{b} % binary variables
\newcommand{\cvar}{\alpha} % continuous variables

\newcommand{\fnP}{\text{SP}}
\newcommand{\fnC}{\text{SC}}
\newcommand{\fnIC}{\text{IC}}

\newcommand{\setfnC}{{\cal SC}}
\newcommand{\setfnIC}{{\cal IC}}

\newcommand{\FPS}{\text{FPS}}

\newcommand{\motor}{\text{m}}
\renewcommand{\frame}{\text{f}}
\newcommand{\sensor}{\text{s}}
\newcommand{\camera}{\text{s}}
\newcommand{\computer}{\text{c}}
\newcommand{\battery}{\text{b}}
\newcommand{\BLP}{BLP\xspace}

\newcommand{\longversion}[2]{#1}

\title{\huge{Robot Co-design: Beyond the Monotone Case}}
\author{Luca Carlone \and Carlo Pinciroli}
\author{Luca Carlone
\thanks{L.\,Carlone is with the Laboratory for
Information \& Decision Systems (LIDS), Massachusetts Institute of Technology, USA,
{\sf lcarlone@mit.edu}}
\and
Carlo Pinciroli
\thanks{C.\,Pinciroli is with Robotics Engineering and the Department of Computer Science, Worcester Polytechnic Institute, USA,
{\sf cpinciroli@wpi.edu}}
\thanks{This work was partially funded by ARL DCIST CRA W911NF-17-2-0181.} \longversion{}{\vspace{-5mm}}
}

\begin{document}
\maketitle

\begin{abstract}
  Recent advances in 3D printing and manufacturing of miniaturized robotic hardware and computing
  are paving the way to build inexpensive and disposable robots. This will have a large impact on
  several applications including scientific discovery (e.g., hurricane monitoring),
  search-and-rescue (e.g., operation in confined spaces), and entertainment (e.g., nano drones).
  The need for inexpensive and task-specific robots clashes with the current practice, where human
  experts are in charge of designing hardware and software aspects of the robotic platform. This
  makes the robot design process expensive and time consuming, and ultimately unsuitable for
  small-volumes low-cost applications. This paper considers the \emph{computational robot co-design}
  problem, which aims to create an automatic algorithm that selects the best robotic modules
  (sensing, actuation, computing) in order to maximize the performance on a task, while satisfying
  given specifications (e.g., maximum cost of the resulting design). We propose a binary
  optimization formulation of the co-design problem and show that such formulation generalizes
  previous work based on strong modeling assumptions. We show that the proposed formulation can
  solve relatively large co-design problems in seconds and with minimal human intervention. We
  demonstrate the proposed approach in two applications: the co-design of an autonomous drone racing
  platform and the co-design of a multi-robot system.
\end{abstract}


\section{Introduction}
\label{sec:intro}

Recent advances in sensor design and rapid prototyping are enabling the manufacturing of low-cost
robots with advanced sensing and perception capabilities. For instance, one can implement
high-precision visual-inertial navigation with inexpensive camera and MEMS inertial measurement
units~\cite{Zhang17rss-vioChip}.  Similarly, modern embedded CPU-GPU~\cite{TX2website} offer
high-performance computing in a compact form-factor and at a relatively affordable cost. These
trends, together with the availability of low-cost micro-motors are enabling fast and cheap
production of robotic platforms. In these cases, the most expensive resource becomes the effort of
the expert human designers who are in charge of designing all the aspects of the robotic platform,
including hardware and software. While this solution is still acceptable in the case where low-cost
robots must be produced in volumes (e.g., vacuum cleaning robots), it may not be desirable when only
few robots need to be deployed. Consider, for instance, the design of a drone for hurricane
monitoring~\cite{Lipinski14aiaa}: the drone must be disposable, hence inexpensive, and it is not
typically produced in volumes. In other contexts, one may need to devise a design quickly, in order
to create a robot to be deployed in a time-sensitive mission. For instance, one may need to design a
search-and-rescue robot tailored to a specific mission (e.g., search for survivors in a narrow cave
with a given size of the entry point). Finally, human design does not necessarily lead to optimal
solutions. Designers typically consider different robotics modules in isolation in order to tame the
design complexity, and such decoupling usually leads to suboptimal performance.

%% OUR CONTRIBUTION
These reasons motivate us to investigate \emph{computational robot co-design}, which aims to create
an automatic algorithm that selects the best robotic modules (sensing, actuation, computing) in
order to maximize the performance on a task, while satisfying given specifications (e.g., maximum
cost of the resulting design). Here the term ``computational'' refers to the fact that the design
techniques can be implemented on a machine, and require minimal human intervention. Moreover, the
term ``co-design'' refers to the attempt to consider the robotic system as a whole, rather than
(sub-optimally) decoupling the design of each module.
% (e.g., computation, sensing, and actuation). 
% In particular, our goal is to push the boundary of robot co-design to tackle real-world 
% robot design problems, building on the recent works~\cite{censi}.

%%% Local Variables:
%%% mode: latex
%%% TeX-master: "../main"
%%% End:

\longversion{
\myParagraph{Related Work} The problem of co-design in robotics touches a wide span of research
topics. In the most general sense, it can be considered as the problem of designing both the mind
(software) and the body (hardware) of the robot at the same time. A few seminal works tackle this
problem from an evolutionary
standpoint~\cite{lipson_automatic_2000,lipson_difficulty_2016,cheney_scalable_2018}, with approaches
that aim to either optimize specific behaviors (e.g., walking) or to explore catalogs of possible
solutions to a general problem (e.g., locomotion). While related to the problem of computational
co-design as discussed above, these approaches typically abstract away several
practical %feasibility
aspects of hardware design, such as power consumption and material selection.

\begin{figure}[t]
\begin{center} % ,natwidth=610,natheight=642
    \includegraphics[width=1\columnwidth,trim=0mm 25mm 0mm 25mm,clip]{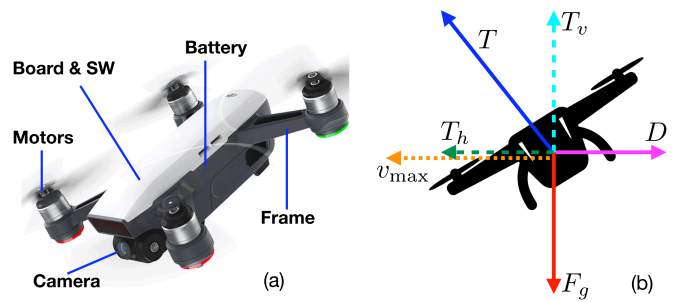}\vspace{-0.6cm}
    \caption{(a) We apply the proposed co-design framework to the design of the modules (motors, frame, computing, camera, battery pack) of
    an autonomous racing drone, (ii) Force diagram during forward motion (side view). \label{fig:drone} } 
\end{center}
\vspace{-0.6cm}
\end{figure}

At the opposite side of the spectrum, a large body of work on software/hardware design exists in the
embedded systems community, e.g.\ FPGA and ASIC design~\cite{Zhang17rss-vioChip}. In this domain,
the focus is on the synthesis of practical solutions that explicitly consider low-level aspects such
as power optimization and real-time scheduling in electronic devices. Since
Chinook~\cite{chou_chinook_1995}, a highly influential early approach, the field of
software/hardware co-design has flourished, reaching relevant applications in safety-critical fields
such as the aerospace~\cite{schafer_divide_2012} and automotive industry. The inception
of domain-specific languages such as the \emph{Architecture Analysis and Design Language}
(AADL)~\cite{noauthor_architecture_nodate} has enabled a systems-of-systems approach in which
co-design can be coupled with a wide array of tools for property verification, both at the software
and hardware level. These tools, however, are thought to support \emph{human} design.
%do not support computational
%are not designed for automatic optimization.

The field of modular robotics attempts to bridge low-level embedded system design with high-level
functionality specifications.
% can be considered as a synthesis between
% low-level, generic, embedded system design, and the need to bridge the
% gap between functionality and control.} 
In modular robotics, the challenge is to create a diverse set of composable, programmable modules
that can be used to form robots with different capabilities. The recent, rapid development of 3D
printing is offering increasingly powerful tools to streamline the automatic design of robotic
platforms, and will arguably foster the development of modular robotics. In a seminal paper, Mehta
\emph{et al.}~\cite{mehta_integrated_2015} demonstrate the feasibility of a procedure that takes a
high-level human-defined specification of a robot and outputs the 3D description files of the
components, along with instructions to assemble the robot, manufacture the electronics, and
automatically generate the control firmware. To the best of our knowledge, no work exists in the
automatic generation of specifications that can be input to this system.

A core issue in computational co-design is the conception of a suitable formalism to express the
design problem. Ideally, an effective formalism for automatic design should combine adequate
granularity along with guarantees of correctness. Evolutionary
methods~\cite{lipson_automatic_2000,lipson_difficulty_2016,cheney_scalable_2018} combine a low-level
parametric representation, such as rigid bodies connected by joints or voxel-based artifacts, with
neural network-based control. The main issue of this representation is the amount of detail involved
in the optimization process, which limits scalability, and the absence of correctness
guarantees. Other formalisms closer to the human design process, such as the one from Mehta \emph{et
  al.}~\cite{mehta_integrated_2015} and AADL~\cite{noauthor_architecture_nodate}, achieve expressive
power by naturally incorporating modularity. Both languages represent a co-design problem as a tree
composed of heterogenous nodes which represent hardware and software aspects of a robot. AADL, in
particular, offers a set of tools specifically designed for early analysis of candidate solutions,
but without support for automatic exploration of the solution space. Correct-by-design compositional
approaches have also been the focus of recent studies in control theory. For instance, Hussien
\emph{et al.}~\cite{hussien_abstracting_2017} proposed an automatic method to decompose a large
control problem into a cascade of simpler problem, under the assumption that the latter are feedback
linearizable, while Kim \emph{et al.}~\cite{kim_constructing_2018} devised a method to arbitrarily
construct system abstractions from simpler, well-posed components.  Lastly,
Censi~\cite{censi_mathematical_2015,censi_class_2017} proposed a powerful co-design approach that
enables the automatic generation of solutions from a specification that includes recursive
constraints. Censi's approach assumes that a monotone mapping between resources and performance
exists, thus casting computational co-design as a fixed-point problem in the
solution space.
%the solution of the design optimization problem as a fixed-point problem in the
% solution space.
}{\input{relatedwork_short}}
\longversion{
%%%%%%%%%%%%%%%%%%%%%%%%%%%%%%%%%%%%%%%%%%%%%%%%%%%%%%%%%%%%
\myParagraph{Contribution} This paper builds on recent work by
Censi~\cite{censi_mathematical_2015,censi_class_2017} and pushes the boundary of computational
co-design by relaxing its monotonicity assumption.  Monotonicity implies that investing more
resources leads to an increase in the performance. However, it is easy to find interesting examples
for which this property does not hold. For instance, in the design if a drone frame, the fact that
we increase the size of the frame (resource) does not imply that the drone will fly faster
(performance). Similarly, in a collaborative transportation problem, deploying a larger number of
robots (resources) does not necessarily lead to an increase in the overall performance of the
system~\cite{hamann_towards_2012}.
% Moreover, we observed that the split between resources and performance, as proposed by Censi, is often 

This paper presents a more general computational co-design approach.  Our problem formulation is
presented in Sec.~\ref{sec:problemFormulation}, where we discuss the characterization of the
design space and classify the design specifications in terms of \emph{system-level performance},
\emph{system-level constraints}, and \emph{intrinsic constraints}.  Sec.~\ref{sec:formulation}
rephrases the resulting co-design problem in terms of a binary optimization problem, where each
binary variable indicates whether a given component is chosen as part of the design or
not. Sec.~\ref{sec:algorithms} discusses in which case we can expect to be able to solve the
binary optimization problem using off-the-shelf optimization tools; in general, binary optimization
is intractable, but there exist several algorithms and implementations that are able to solve
moderate-sized problems efficiently as long as some property (e.g., linearity) holds.  We show that
our approach allows recasting several functions that are nonlinear in the properties (or
\emph{features}) of the robotic modules as linear functions.  The proposed optimization framework
also makes it possible to model other constraints found in practice, such as \emph{compatibility}
constraints (e.g., we cannot use a LIDAR-based algorithm to process data from a monocular camera),
which was beyond reach for existing methods~\cite{censi_mathematical_2015}.

% and can be extended to include continuous variables; both these aspects are 
% Our overarching goal is to isolate a large set of co-design problems 
% \begin{itemize}
% \item going beyond monotone design
% \item list of innovations wrt SOTA:
% \item (online algorithms? large scale problems?)
% 	\begin{itemize}
% 	\item Censi's complexity still grows exponentially (?)
% 	\item test a large instance on Censi's sw and show that it is very slow
% 	\end{itemize}
% \item compatibility constraints: e.g., choice of sensors influences/restricts choice of algorithms
% \item (continuous variables)
% \item interconnections design
% \end{itemize}
% allow leveraging tools and theory from numerical optimization.
%
% Intuitive problem statement & motivations
% motivations: multi robot design, robot design

We conclude the paper by presenting two applications of the proposed co-design approach to the
design of an autonomous drone racing platform and a multi-robot system.  These examples are
presented in Sec.~\ref{sec:experiments}.  Drone design is an interesting topic per se, and has
already received attention in the literature~\cite{Kushleyev13auro-MAVswarms,Zhang17rss-vioChip},
which however still lacks a computational design approach.  New and untapped challenges arise in the
co-design of multi-robot systems.  We are not aware of approaches that specifically tackle this
problem; however, a vast literature exist on the automatic generation of control systems for robots
swarms, including evolutionary methods~\cite{trianni_evolutionary_2008} and automatic generation of
composable control structures~\cite{francesca_automode-chocolate:_2015}.

%%% Local Variables:
%%% mode: latex
%%% TeX-master: "../main"
%%% End:
}{\input{contribution_short}}

\section{General Robot Co-design}
\label{sec:problemFormulation}

The co-design problem consists in jointly designing robot software and hardware components to
maximize task-dependent performance metrics (e.g., endurance, agility, payload) subject to
constraints on the available resources (e.g., cost). The complexity of the problems stems from the
fact that a robotic system involves an intertwining of modules. Each module contributes to the
overall performance of the system while potentially consuming resources. In this paper, we consider
the realistic case in which we have to choose each module (e.g., motor, embedded computer, camera)
in the robotic system from a given catalog, and we formulate the co-design as the combinatorial
selection of modules that maximize performance while satisfying given system-level and module-level
constraints.
 % and for each module we have to choose the most suitable module design (e.g., a s)
 %  which the design of each module (e.g., the actuation module) 
 % involves both ``discrete'' decisions (e.g., which brand of motors to choose) and the tuning of ``continuous'' parameters 
 % (e.g., gains of the control loops). Hence, we provide a general formulation to co-design discrete and continuous variables 
 % defining the overall robotic system.
 % This section provides our formulation of the co-design problem. 
 % In particular, 
In Sec.~\ref{sec:modules} we introduce our abstraction of the modules forming the overall robotic
system. In Sec.~\ref{sec:interactions} we describe the interactions among the modules and how they
contribute to the overall system performance and resource
utilization. %Finally, Sec.~\ref{sec:formulation}.
In Sec.~\ref{sec:formulation} we state the co-design problem as a binary optimization problem.

%  More specifically, the  
% is the joint design of the modules of an interconnected system given specifications on the desired performance and the available resources the system can use.  
% % Start introducing notation
% modules: design space:  
% \begin{figure}[t]
% \begin{center}
%     \includegraphics[width=0.8\columnwidth]{architecture}
%     \caption{Robotic System.}
%     \label{fig:architecture}
% \end{center}
% \end{figure}

%%% Local Variables:
%%% mode: latex
%%% TeX-master: "../main"
%%% End:

%%%%%%%%%%%%%%%%%%%%%%%%%%%%%%%%%%%%%%%%%%%%%%%%%%%%%%%%%%%%%%%%%%%%%%%%
\subsection{Modules, Catalogs, Features}
\label{sec:modules}

We consider the case in which the robotic system comprises a given set of \emph{modules} $\calM$.
The modules may include, for instance, the actuators, the sensors, the computational board, the
perception algorithms, the control algorithms, the planning algorithms, etc.
% Let us consider the reference architecture of a robotic system, shown in \Fig{fig:architecture}. 
% The architecture is comprised of a set of \emph{modules}, divided in \emph{physical} layer (sensing and actuation modules), 
% and \emph{computational} layer (e.g., estimation, perception, control, and planning modules).
%  As we will see the notion of ``module'' is more general and does not need to be constrained to the architecture in  
%  \Fig{fig:architecture} (for instance, we can easily accommodate learned modules replacing one of more modules in \Fig{fig:architecture}).
%   However, we keep \Fig{fig:architecture} as running example for the reader's convenience.
For each module $i \in \calM$, we have a \emph{catalog} $\calC_i$ of potential choices: for
instance, we can purchase different motor models, or we can utilize different approaches and
implementations of a planning algorithm.

\myParagraph{Design vector} The goal of the co-design is to select an element $j \in \calC_i$
(catalog for module $i$), for each module $i \in \calM$.  We can represent this selection using a
binary vector $\vxx_i$ for each module, where the $j$-th entry of $\vxx_i$ is 1 if we select the
$j$-th element in the catalog $\calC_i$ or zero otherwise. Clearly,
$\vxx_i \in \{0,1\}^{|\calC_i|}$, where $|\calC_i|$ is the cardinality (number of elements) of the
catalog. Therefore, the design is fully specified by the \emph{design vector} $\vxx$ obtained by
stacking $\vxx_i$ for each $i \in \calM$.  The design vector has size
$N \doteq \sum_{i\in\calM} |\calC_i|$, which we refer to as the \emph{dimension of the design
  space}.

\myParagraph{Feature matrix} Each module has a number of \emph{features} describing the technical
specifications of the module.  For example, the features of a motor may include the cost, torque,
weight, maximum speed, power consumption, size of the motor, among other technical data.  In
general, the features are a list of properties one would find in the datasheet of a component.
Similarly, for an algorithm, the set of features may include information about the expected
performance and computational cost of the algorithm.
% While the descriptor characterizes the design choices for a single module, for the design purposes it is convenient 
% to associate these choices to a set of technical specifications. For instance, while setting $\dvar_1 = 2$ may mean that we
%  selected the second motor in our catalog, our computational co-design approach requires defining what this choice of motor 
%  entails in terms of cost, weight, power consumption, etc. 
%  Therefore, we define the notion of \emph{feature}, which is the (potentially large) set of specifications associated with 
%  a module descriptor. Mathematically, the \emph{feature} vector for module $i$ is a function of the descriptor: 
%  $\fvec\at{i} = \fnF\at{i}(\mvec\at{i})$, which maps the module design choices to the corresponding technical specifications.
%  Intuitively, the feature vector can contain all the parameters listed in the datasheet of a component.

Clearly, each element in the catalog of module $i$ (e.g., different motor models) will have
different values of each feature.  We can thus succinctly describe the list of features for each
element in the catalog of module $i$ as a \emph{feature matrix} $\MF_i$, where each row correspond
to a given feature, and different columns correspond to different elements in the catalog.  For
instance, in a toy problem, we can have the following feature matrix for the motor module:

\vspace{-3mm}
\small
\bea
% \text{motor 1} \text{motor 2} \text{motor 3}  \nonumber \\ 
\MF_{\motor} = 
\left[
\begin{array}{ccc}
W\at{1} & W\at{2} & W\at{3} \\ 
V\at{1} & V\at{2} & V\at{3} \\ 
A\at{1} & A\at{2} & A\at{3} \\ 
C\at{1} & C\at{2} & C\at{3} \\ 
T\at{1} & T\at{2} & T\at{3} 
\end{array} 
\right]
\begin{array}{l}
\leftarrow \text{weight}\\
\leftarrow \text{voltage}\\
\leftarrow \text{current}\\
\leftarrow \text{cost}\\
\leftarrow \text{torque}
\end{array}
\label{eq:featureTable_example}
\eea

\normalsize
%2:weight(g) 3:voltage(V)  4:current(A)  5:cost($)  6:thrust(g) 
\noindent where the $j$-th column describes the features of the
$j$ element in the motor catalog. In practice, the values in the matrix are known from datasheet or
from prior experiments.
% Therefore it holds:
% \beq
% \vf_i = \MF_i \; \vxx_i 
% \eeq

\begin{remark}[Beyond monotonicity]
  A way to relate our approach (describing modules via catalogs and feature matrices) to the one of
  Censi~\cite{censi_mathematical_2015} is as follows. Censi splits what we call \emph{features} into
  resources and functionalities (imagine for instance, that the first
  $n_r$ rows or~\eqref{eq:featureTable_example} are labeled as ``resources'' and the last $n_p$ as
  ``functionalities''). Then, Censi assumes that the columns of~\eqref{eq:featureTable_example}
  satisfy the monotonicity property, i.e., choices of components (columns) leading to better
  functionalities require more resources.  This is not necessarily true in practice: if we design a
  drone system, choosing a larger frame (i.e., consuming more resources) does not necessarily imply
  that our drone will fly faster (i.e., better functionality). %In general
  We go beyond~\cite{censi_mathematical_2015} with two main innovations. First, the entries in our
  feature matrix~\eqref{eq:featureTable_example} are completely arbitrary, hence relaxing the
  monotonicity assumption\longversion{. We will further highlight the importance of relaxing
    monotonicity for multi-robot design %when discussing the design of a multi-robot system
    in Sec.~\ref{sec:experiments-multiRobot}.}{. } Second, while resources and functionalities of
  each module may be problem-dependent, feature matrices are \emph{design-agnostic}, as discussed
  in~Remark~\ref{rmk:agnostic}\longversion{ below.}{.}
\end{remark}

\begin{remark}[Feature matrices are design-agnostic]
\label{rmk:agnostic}
The notion of ``module'' and ``feature matrix'' are agnostic of the co-design problem, which is only
defined in Sec.~\ref{sec:formulation}. Indeed, the notion of ``features'' is general---the same
modules can be used in any co-design problem involving that module. This is in sharp contrast
with~\cite{censi_mathematical_2015}, where the definition of ``resources'' and ``functionalities''\longversion{ (or ``performance'') for}{ for} 
 each module depends on the interactions between modules, hence it is
problem-dependent.  For instance: Fig.~21 in~\cite{censi_mathematical_2015} classifies the battery
capacity as a performance metric, but in other problems the battery capacity can be a resource.
% \LC{battery example is not superconvincing - do you have an alternative example?}
Our choice to create an intermediate abstraction, the \emph{feature matrix}, resolves this
dependence, enabling re-usability of modules across problems.
% \LC{I do not remember this:
% Indeed, to preserse coherence of the performance/resources interconnections, [Censi] has also to treat some mathematical relations (products, sums) as independent design modules, which appears to be an artifact of his formulation (e.g., Fig. 23). 
% }
\end{remark}

% \red{
% {\bf remark}: Reasoning at different granularities:
% - granularity of modules is completely arbitrary (figure): I can design a system with 3 modules (Fig.(a)) given potential choices for 1,2,3, or design 1-2 in conjunction if I have a list of available designs for the combo 1-2 (Fig.(b)). We can compress 2 modules by simply concatenating the corresponding design vectors, which fully define both modules.
% }

%%% Local Variables:
%%% mode: latex
%%% TeX-master: "../main"
%%% End:

%%%%%%%%%%%%%%%%%%%%%%%%%%%%%%%%%%%%%%%%%%%%%%%%%%%%%%%%%%%%%%%%%%%%%%%%
\subsection{Design Specifications: Performance and Constraints}
\label{sec:interactions}

In a robotic system, the different modules interact to contribute to the overall performance of the
system and potentially consume resources. In particular, both the overall system as well as each
module require a minimum amount of resources to operate properly, thus imposing \emph{constraints}
on the design.

In this context, we distinguish three main \emph{design specifications}: \emph{system-level
  performance}, \emph{system-level constraints}, and \emph{module-level constraints}. Intuitively,
system-level performance defines a set of metrics the co-design has to maximize, while system-level
(resp. module-level) constraints specify a set of constraints that need to be satisfied for the
overall system (resp. each module) to operate correctly. These design specifications, which we
discuss in more detail below, together with the catalogs of the modules we want to design, fully
specify the co-design problem (Section~\ref{sec:formulation}).

% ============================================
\myParagraph{Explicit specifications: system-level performance and constraints} As an input to the
co-design process, the user provides a set of performance metrics the design must maximize, as well
as a set of constraints the \emph{overall system} must satisfy. These specifications, which we refer
to as \emph{explicit specifications}, are at the system level, in the sense that they describe the
task that the system must perform.

\emph{System-level Performance.} The \emph{system-level performance} is a vector-value function of a
choice of components.  Recalling that a design is fully characterized by the design vector $\vxx$,
the system-level performance is a function:
\beq
\fnP(\vxx) : \{0,1\}^N \mapsto \Real{N_p},
\label{eq:sp}
\eeq
where $N$ is the dimension of the design space %($N \doteq \sum_{i\in\calM} |\calC_i|$)
and $N_p$ is the number of performance metrics the user specifies.  To clarify~\eqref{eq:sp}, let
us consider a simple example, in which the design must maximize the torque of the wheel motors of an
autonomous race car. In this case, the system-level performance is described by:
\beq
\fnP(\vxx) = 4 \; [\MF_\motor]_T \; \vxx_\motor % = 4 \; \ve_T\tran \; \MF_\motor \; \vxx_\motor
\label{eq:sp_example}
\eeq
In~\eqref{eq:sp_example}, $\MF_\motor$ is the motor feature matrix
in~\eqref{eq:featureTable_example}, $\vxx_\motor$ is the design vector for the motor module, and
$[\MF_\motor]_T$ extracts the row of $\MF_\motor$ corresponding to the motor torque. The factor
``4'' captures the fact that, for simplicity, we assumed four wheels mounting identical motors.
% $\ve_T$ is a vector of size equal to the number of moto which is zero everywhere, 
% except the entry corresponding to the motor 
We remark that the (linear) operator $[\cdot]_T$ selects the row corresponding to a specific feature
from the feature matrix, while the multiplication by $\vxx_\motor$ has the effect of selecting a
single column (i.e., choosing a motor model) from $\MF_\motor$, due to the fact that $\vxx_\motor$
has a single non-zero element.

\emph{System-level Constraints.} The \emph{system-level constraints} are (scalar) equality or
inequality constraints, describing hard requirements on the desired behavior of the system or
constraints on the resources that can be used for the design.  We express system-level constraints
involving the design vector $\vxx$: % together with a relation:
\bea
% \left\{ \fnC(\vxx)_k : \{0,1\}^N \mapsto \Real{N_{sck}} \quad , \quad \text{relation} \in \{=,\leq\}\right\}
\fnC(\vxx)_k \leq 0 & \text{for } k \in \setfnC_{\leq}   \\
% \quad \text{or} \quad 
\fnC(\vxx)_{k'} = 0 & \text{for } k' \in \setfnC_{=} 
\label{eq:sc}
\eea
where $\setfnC_{\leq}$ and $\setfnC_{=}$ are the sets of inequality and equality system-level
constraints, respectively. For instance, we can have an upper bound on the overall cost of the
design:
\beq
\sum_{i\in\calM} \; [\MF_i]_C \; \vxx_i \leq \text{budget} \Rightarrow \fnC(\vxx)_k = \sum_{i\in\calM} \; [\MF_i]_C \; \vxx_i - \text{budget} % = 4 \; \ve_T\tran \; \MF_\motor \; \vxx_\motor
\label{eq:sc_ineq_example}
\eeq
where, as before, the linear operator $[\cdot]_C$ extracts the cost from the feature matrix of each
module $i$.  Note that the ``budget'' must be provided by the user and it is specific to the design
instance, hence~\eqref{eq:sc_ineq_example} is a system-level constraint.

An example of system-level equality constraint is the case in which the user wants to consider only
a subset of elements in the catalog (e.g., within a catalog of motors, only two motors are available
in-house). For instance, for the choice of module $i$ to be restricted to the subset
$\calS_i \subset \calC_i$, the user can add the following system-level equality constraint:
\beq
\longversion{}{\textstyle}\sum_{j\in\calS_i} [\vxx_i]_j = 1 \quad\Rightarrow \quad \fnC(\vxx)_{k'} = \sum_{j\in\calS_i}[\vxx_i]_j - 1 % = 4 \; \ve_T\tran \; \MF_\motor \; \vxx_\motor
\label{eq:sc_eq_example}
\eeq
where again $[\vxx_i]_j$ selects the $j$-th element of $\vxx_i$; the
constraint~\eqref{eq:sc_eq_example} enforces one element in $\calS_i$ to be chosen, due to the
binary nature of the vector $\vxx_i$.

% ============================================
\myParagraph{Implicit specifications: module-level constraints} The user must provide explicit
specifications to describe \emph{what} the robotic system is required to do within \emph{which}
operational constraints. On the other hand, \emph{implicit} specifications are transparent (or
uninteresting) to the user and are only needed to guarantee that each module has \longversion{sufficient
resources to function as expected.}{enough resources to function.}
% This are usually design-independent and can be tough 
% In order to operate properly, all modules in the system need to be provided suitable resources (e.g., power, voltage, form-factor).
% These resources (at the module level) are not necessarily of interest for the user, but are still necessary for the operation of the 
% overall system. 

Similarly to the system-level performance, we express module-level (implicit) constraints as
inequality or equality constraints involving the design vector $\vxx$:
\bea
% \left\{ \fnC(\vxx)_k : \{0,1\}^N \mapsto \Real{N_{sck}} \quad , \quad \text{relation} \in \{=,\leq\}\right\}
\fnIC(\vxx)_k \leq 0 & \text{for } k \in \setfnIC_{\leq}   \\
% \quad \text{or} \quad 
\fnIC(\vxx)_{k'} = 0 & \text{for } k' \in \setfnIC_{=} 
\label{eq:sc}
\eea
where $\setfnIC_{\leq}$ and $\setfnIC_{=}$ are the sets of inequality and equality module-level
constraints, respectively.  While the mathematical nature of the system-level and module-level
constraints is similar, we believe it makes sense to distinguish them, since the user has control
over system-level constraints (e.g., to increase the design budget in~\eqref{eq:sc_ineq_example}),
while he/she typically does not have control over the implicit constraints.

For instance, the user cannot change the fact that, for the system to function properly, the onboard
battery ``\battery'' has to provide enough power for all the active modules (say, motors ``\motor'',
sensors ``\camera'', and computing ``\computer''), which indeed translates into an implicit
inequality constraint:
\beal
\hspace{-0cm} A_\battery V_\battery \geq A_\motor V_\motor +  A_\camera V_\camera + A_\computer V_\computer \quad\Rightarrow   \!\!\!\!\\ 
\quad \fnIC(\vxx)_{k}  = \!\!\!\! \displaystyle\sum_{i=\{\motor,\camera,\computer\}} \!\!\!\!
([\MF_i]_A \vxx_i) ([\MF_i]_V \vxx_i)  - ([\MF_b]_A \vxx_b) ([\MF_\battery]_V \vxx_\battery) \!\!\!\!
\label{eq:ic_ineq_example}
\eeal
where $A_i$ and $V_i$ are the current and voltage at module $i$, and $\MF_i$ is again the feature
matrix for module $i$.

Implicit constraints are also useful to model \emph{compatibility constraints}, which, again, the
user is not typically free to alter.  For instance, we may want to model the fact that we cannot run
a certain algorithm (e.g., designed for FPGA) on a certain hardware (e.g., CPU), or we cannot use a
LIDAR-based signal processing front-end to process data from a monocular camera.  Compatibility
constraints can be expressed as follows, for a pair of modules ``a'' and ``b'':
\bea
[\vxx_a]_j \leq  \longversion{}{\textstyle}\sum_{j' \in \calS_j} [\vxx_b]_{j'} &  \text{(compatibility)} \nonumber \\
\;[\vxx_a]_j \leq 1 - \longversion{}{\textstyle}\sum_{j' \in \calS_j} [\vxx_b]_{j'} &  \text{(incompatibility)}
\label{eq:ic_eq_example}
\eea
The first inequality in~\eqref{eq:ic_eq_example} imposes that, when the $j$-th element in the
catalog of module ``a'' is selected, we can only choose the module $b$ from the subset $\calS_j$
(subset of \emph{compatible} modules in the catalog of $b$). The second inequality
in~\eqref{eq:ic_eq_example} imposes that, when the $j$-th element in the catalog of module ``a'' is
selected, then we cannot choose the module $b$ from the subset $\calS_j$ (subset of
\emph{incompatible} modules in the catalog).

%%% Local Variables:
%%% mode: latex
%%% TeX-master: "../main"
%%% End:

%%%%%%%%%%%%%%%%%%%%%%%%%%%%%%%%%%%%%%%%%%%%%%%%%%%%%%%%%%%%%%%%%%%%%%%%
\subsection{Co-design}
\label{sec:formulation}

The co-design problem can be now stated as follows.

\begin{definition}[Robot Co-design]
  Given the catalogs (and the corresponding feature matrices $\MF_i$) for each module $i \in \calM$
  to be designed, given the system-level performance function $\fnP(\vxx)$ and the set of
  system-level inequality and equality constraints ($\setfnC_{\leq},\setfnC_{=}$), as well as the
  set of module-level inequality and equality constraints ($\setfnIC_{\leq},\setfnIC_{=}$), robot
  co-design searches for the choice of modules $\calM$ that maximizes the system-level performance,
  while satisfying the constraints:
  \bea
  \label{eq:genCodesign}
  \max_{\vxx \in\calX} & \fnP(\vxx)  \\
  \subject & \fnC(\vxx)_k \leq 0 & \text{for } k \in \setfnC_{\leq} \nonumber  \\
           & \fnC(\vxx)_{k'} = 0 & \text{for } k' \in \setfnC_{=}  \nonumber\\
           & \fnIC(\vxx)_k \leq 0 & \text{for } k \in \setfnIC_{\leq} \nonumber\\
           & \fnIC(\vxx)_{k'} = 0 & \text{for } k' \in \setfnIC_{=}  \nonumber
  % \\ & \sum_j [x_i]_j = 1  & \text{for } i \in \calM 
  \eea
  where $\calX$ is the set of binary vectors that correspond to unique choices of each module
  (mathematically:
  $\calX \doteq \setdef{ \{0,1\}^N }{ \sum_{j=1}^{|\calC_i|} [\vxx_i]_j = 1 \text{ for each module }
    i \in \calM}$).
\end{definition}

Problem~\eqref{eq:genCodesign} is a \emph{binary} optimization problem, since the vector-variable
$\vxx$ has binary entries.  The formulation does not take any assumption on the nature of the
functions involved in the objective function and the constraints. Our formulation is this
\emph{general} and it \emph{does not assume monotonicity}: indeed, by introducing the notion of
``feature matrix'', we circumvented the problem of reasoning in terms of resources and functionality
of each module. As we will see, this framework allows one to solve non-monotonic problems (see
Sec.~\ref{sec:experiments}).

\begin{remark}[Total Order]
  In general, the optimization problem~\eqref{eq:genCodesign} is a multi-objective maximization,
  since the objective is vector-valued.  While a major concern in~\cite{censi_mathematical_2015} was
  how to deal with partially ordered sets (e.g., vectors), here we take a more pragmatic
  approach. In the formulation~\eqref{eq:genCodesign} we assume a total order for the vector in the
  objective, while we restricted the constraints to be scalar equalities and inequalities, hence
  working on the totally ordered set of reals. In particular, we use the \emph{lexicographical
    order} to enforce a total order on a vector space of performance vectors.  In the
  lexicographical order two vectors $\vy = [y_1\;y_2\;\ldots\;y_n]$ and
  $\vz = [z_1\;z_2\;\ldots\;z_n]$ satisfy $\vy \leq \vz$ if an only if $y_1 \leq z_1$, or $y_1=z_1$
  and $y_2 \leq z_2$, or $y_1=z_1$, $y_2=z_2$ and $y_3 \leq z_3$ etc.  This order implies that the
  entries of the vector are sorted by ``importance''. For instance, if we \emph{minimize} a
  performance vector that includes $[\text{cost}, \; \text{size}]$, then we search for the design
  that minimizes cost, and if two designs have the same cost, we prefer the design with smaller
  size.  Note that we can also use the lexicographical order to generalize the constraints to be
  vector-valued functions.
\end{remark}

\section{Linear Co-design Solvers}
\label{sec:algorithms}

We now discuss several cases in which we can expect to solve Problem~\eqref{eq:genCodesign} in
reasonable time using off-the-shelf optimization tools.  While Problem~\eqref{eq:genCodesign} is
fairly general (we did not take any assumptions on the functions involved in the objective and the
constraints), we do not expect to be able to solve~\eqref{eq:genCodesign} globally and efficiently
in general. Indeed, binary optimization is NP-hard and the computational cost of solving a problem
grows exponentially with its size~\cite{Schrijver86book-integerProgramming}.

Despite the intrinsic intractability of binary optimization, integer and binary programming
algorithms keep improving and modern implementations (e.g., IBM CPLEX~\cite{CPLEXwebsite}) are
already able to solve \emph{linear and quadratic} binary optimization problems involving thousands
of variables in reasonable time (i.e., seconds to few minutes).  In our co-design problem, this
means that we can expect to solve problems with $N = 2,000$ (dimension of $\vxx$) in seconds, which
would be the case if we have to design $10$ modules, where each module catalog has $200$ potential
choices (remember $N \doteq \sum_{i\in\calM} |\calC_i|$).

The possibility of solving linear and quadratic binary optimization problems of interesting size in
reasonable time motivates use to investigate when we can expect to rephrase~\eqref{eq:genCodesign}
as a linear or quadratic optimization problem (note: the answer is not as trivial as it might seem).
In the rest of this section we focus on the cases where we can rephrase~\eqref{eq:genCodesign} as a
binary \emph{linear} program (\BLP), since this already includes several cases of practical
interest.
% and we are now able to solve 
% some classes of problems in reasonable time (i.e., few seconds to minutes) for relatively large instances (i.e., )
% there 
% exist several algorithms and implementations (e.g., CPLEX~\cite{CPLEXwebsite}) 
% that are able to solve efficiently moderate-sized problems as long as some property (e.g., linearity) holds. 
% Note: we can model every connection if we are willing to increase problem size (include produce spaces), otherwise we can keep original dimension as long as we assume convexity.
Since the linearity of Problem~\eqref{eq:genCodesign} relies on the capability of expressing both
the objective and the constraints are linear functions, in the following we discuss which type of
functions we can expect to rephrase as linear.

% ==========================================================
\myParagraph{(a) Linear functions} Choosing linear functions in the objective and constraints would
make~\eqref{eq:genCodesign} a \BLP. Therefore, if the objective and the constraints
in~\eqref{eq:genCodesign} have the following form, then~\eqref{eq:genCodesign} is a \BLP:
\beq
f(\vxx) = \longversion{}{\textstyle}\sum_{i \in \calM} \vv_i\tran \vxx_i + \text{constant}
\label{eq:f_linear}
\eeq
where $\vv_i$ is a known vector. We remark that most of the examples in
Section~\ref{sec:interactions},
including~\eqref{eq:sp_example},~\eqref{eq:sc_ineq_example},~\eqref{eq:sc_eq_example},
and~\eqref{eq:ic_eq_example}, can be directly expressed in this form.

% ==========================================================
\myParagraph{(b) Sum of nonlinear functions of a module}
Under the setup of Section~\ref{sec:problemFormulation}, we can express the sum of any nonlinear
function involving a single module as a linear function. Consider for instance the following
function:
\beq f(\vxx) = \longversion{}{\textstyle}\sum_{i \in \calM} f_i(\MF_i \vxx_i)
\label{eq:f_sumnonlinear1}
\eeq
where each function $f_i$ depends nonlinearly on the features of module $i$.  We already observed
that $\MF_i \vxx_i$ picks the $j$-th column of the feature table $\MF_i$, where $j$ is the only
non-zero entry in $\vxx_i$ and is equal to $1$. It is easy to show that:
\beq
f_i(\MF_i \vxx_i) = \hat{f}_i(\MF_i) \vxx_i
\label{eq:trick_lin_nonlin}
\eeq
where $\hat{f}_i(\MF_i)$ denotes the (known) row vector obtained by applying $f_i(\cdot)$
column-wise to the matrix $\MF_i$:
\beq
\hat{f}_i(\MF_i) \doteq 
\left[
\begin{array}{cccc}
f_i(\MF_i\at{1}) & f_i(\MF_i\at{2}) & \ldots & f_i(\MF_i\at{|\calC_i|})
\end{array}
\right]
\eeq
Using~\eqref{eq:trick_lin_nonlin}, we can rewrite~\eqref{eq:f_sumnonlinear1} as:
\beq
f(\vxx) = \longversion{}{\textstyle}\sum_{i \in \calM} \hat{f}_i(\MF_i) \vxx_i
\label{eq:f_sumnonlinear_linearized}
\eeq
where $\hat{f}_i(\MF_i)$ is a known vector. The expression~\eqref{eq:f_sumnonlinear_linearized} is
now linear and has the same form as~\eqref{eq:f_linear}.  Note that the
``trick''~\eqref{eq:trick_lin_nonlin} only holds since in our co-design problem each $\vxx_i$ has a
single non-zero entry equal to 1.  As an example, \eqref{eq:ic_ineq_example} involves a sum of
nonlinear functions of each module and so, according to our discussion, it can be expressed as a
linear constraint.

% ==========================================================
\myParagraph{(c) Rational functions} We now show that we can express a family of rational functions
of the modules as linear functions.  Let us start by considering the case in which the objective
function in~\eqref{eq:genCodesign} is a rational function in the following form:
\beq
f(\vxx) = \frac{ \prod_{i\in\calN} f_i(\MF_i \vxx_i) }{ \prod_{i\in\calD} f_i(\MF_i \vxx_i) }
\label{eq:f_sumnonlinear2}
\eeq
where $\calN \subseteq \calM$ and $\calD \subseteq \calM$ are arbitrary subsets of modules.
Interestingly, it is possible to transform~\eqref{eq:f_sumnonlinear2} into a linear function.  For
this purpose, we note that maximizing a quantity is the same as maximizing its logarithm, since the
logarithm is a non-decreasing function. Therefore, we can replace the objective
function~\eqref{eq:f_sumnonlinear2} with the equivalent objective:
\beq
\!\!f'(\vxx) = \log\frac{ \displaystyle\prod_{i\in\calN} f_i(\MF_i \vxx_i) }{ \displaystyle\prod_{i\in\calD} f_i(\MF_i \vxx_i) } \!\!= 
\!\!\sum_{i\in\calN} \log f_i(\MF_i \vxx_i) - \sum_{i\in\calD} \log f_i(\MF_i \vxx_i)
\label{eq:f_sumnonlinear3}
\eeq
Now, we note that \eqref{eq:f_sumnonlinear3} is a sum of nonlinear functions involving a single
module, hence it can be simplified to the following linear function:
\beq
f'(\vxx) = \sum_{i\in\calN} \log \hat{f}_i(\MF_i) \vxx_i - \sum_{i\in\calD} \log \hat{f}_i(\MF_i) \vxx_i
\label{eq:f_sumnonlinear4}
\eeq
The same approach can be applied to inequality (as well as equality) constraints in the following
form ($\bar{r}$ is a given scalar):
\beq
\frac{ \prod_{i\in\calN} f_i(\MF_i \vxx_i) }{ \prod_{i\in\calD} f_i(\MF_i \vxx_i) } \leq \bar{r} 
\label{eq:f_sumnonlinear5}
\eeq
which can be reformulated as equivalent linear constraints.

% ==========================================================
\myParagraph{(d) Nonlinear functions of multiple modules} We conclude this section by considering
the more general case, in which we have a non-linear function involving multiple modules. For
simplicity of exposition, let us consider the case of a generic non-linear function involving two
modules ``a'' and ``b'':
\beq
f(\vxx) = f(\vxx_a , \vxx_b)
\label{eq:f_sumnonlinear_multiple}
\eeq
When $f(\cdot)$ does not have a specific structure (as the cases discussed above), it is still
possible to obtain a linear expression for~\eqref{eq:f_sumnonlinear_multiple}, but, as we will see,
we will pay a price for this lack of structure.  To express~\eqref{eq:f_sumnonlinear_multiple} as a
linear function, we introduce an extra variable, a matrix $\MZ_{ab}$ which has size
$|\calC_a| \times |\calC_b|$ and that represents the joint choice of the modules ``a'' and ``b''. In
other words, $\MZ_{ab}$ is zero everywhere and has a single entry equal to $1$ in row $j_a$ and
column $j_b$, when we choose element $j_a$ for module ``a'' and element $j_b$ for module ``b''.
Clearly, the variables $\MZ_{ab}$, $\vxx_a$, and $\vxx_b$ are not independent and they have to
satisfy the following linear constraints:
\beq
\MZ_{ab} \ones = \vxx_b \quad,\quad \ones\tran \MZ_{ab} = \vxx_a
\eeq
where $\ones$ is a vector of ones of suitable dimension ($\MZ_{ab} \ones$ returns the row-wise sum
of the entries of $\MZ_{ab}$, while $\ones\tran \MZ_{ab}$ returns the column-wise sum). Intuitively,
the constraints make sure that the matrix $\MZ_{ab}$ and the vectors $\vxx_a$ and $\vxx_b$ encode
the same choice of components. By introducing the variable $\MZ_{ab}$, we can
write~\eqref{eq:f_sumnonlinear_multiple} as a linear function of $\MZ_{ab}$ following the same ideas
of case (b) discussed above. The price to pay is an increase in the size of the optimization
problem.  In the general case, in which more than 2 variables are involved in a generic non-linear
function, $\MZ$ becomes a (sparse) tensor with a number of entries equal to
$\prod_{i \in \calN} |\calC_i|$, where $\calN$ is the set of modules involved in the (generic)
non-linear transformation.

When increasing the size of the optimization problem is not an option, we can still
use~\eqref{eq:genCodesign} and substitute non-linear functions with linear (or \emph{linearizable},
as the cases above) surrogates.  We show an example of this approach in
Sec.~\ref{sec:experiments-drone}.

\section{Codesign Experiments}
\label{sec:experiments}

This section presents two examples of applications of the proposed co-design approach:
Sec.~\ref{sec:experiments-drone} focuses on the co-design of an autonomous racing drone, while
Sec.~\ref{sec:experiments-multiRobot} considers the co-design of a team of robots for
collaborative transportation.  In both examples we use IBM CPLEX~\cite{CPLEXwebsite} to solve the
binary optimization program~\eqref{eq:genCodesign}. % that computes the optimal design.

%%% Local Variables:
%%% mode: latex
%%% TeX-master: "../main"
%%% End:

% \longversion{\input{experiments-drone}}{\input{experiments-drone_short}}
\longversion{
\newcommand{\budget}{\$1000\xspace}

% ==========================================================
\subsection{Autonomous Drone Co-design}
\label{sec:experiments-drone}

This section applies the proposed co-design framework to the problem of designing 
an autonomous micro aerial vehicle for drone racing. 
In particular, we answer the question: \emph{what is the best autonomous drone design we can obtain on a \budget budget, 
using components chosen from a given catalog}?

\myparagraph{Modules} 
The first step of the design process is to identify the set of modules $\calM$ we want to design and 
 prepare a catalog for each module.
 In this example, we design five key modules that form an autonomous drone: 
motors, frame, computation, camera, and battery. 
For the motors, frame, and batteries, we selected real components 
 that are commonly used for drone racing. In particular, we considered 17 candidate 
 motors, 6 candidate frames, and 12 candidate batteries. The websites we used to select those components 
 (together with their cost and their specs) as well as the actual numbers we used in this example
  are available at \url{https://bitbucket.org/lucacarlone/codesigncode/}.
 For the camera and the computation, we took a number of simplifying assumptions. 
 For the camera, we disregarded the choice of the lens and we defined a set of (realistic) candidate 
 cameras, where, however, only a single one is chosen from an actual catalog. 
 For the design space of the ``computation'', we actually considered the joint selection of 
 the computing board and the visual-inertial navigation (VIN) algorithm used for state estimation. 
 Considering them as a single component has the advantage of allowing the use of statistics 
 reported in other papers, e.g., stating that a VIN algorithm runs at a given framerate on a given computer~\cite{Zhang17rss-vioChip}.
 Therefore our modules are $\calM \doteq \{\motor, \frame, \camera, \computer, \battery\}$ 
 ($\motor =$ motors, $\frame =$ frame, $\camera =$ camera, $\computer =$ computer, $\battery = $ battery).
 % where $\motor, \frame, \camera, \computer, \battery$ denote the motors, 
 %frame, camera, computer, and battery. 

 \begin{figure}[t]
\begin{center} % ,natwidth=610,natheight=642
    \includegraphics[width=1\columnwidth]{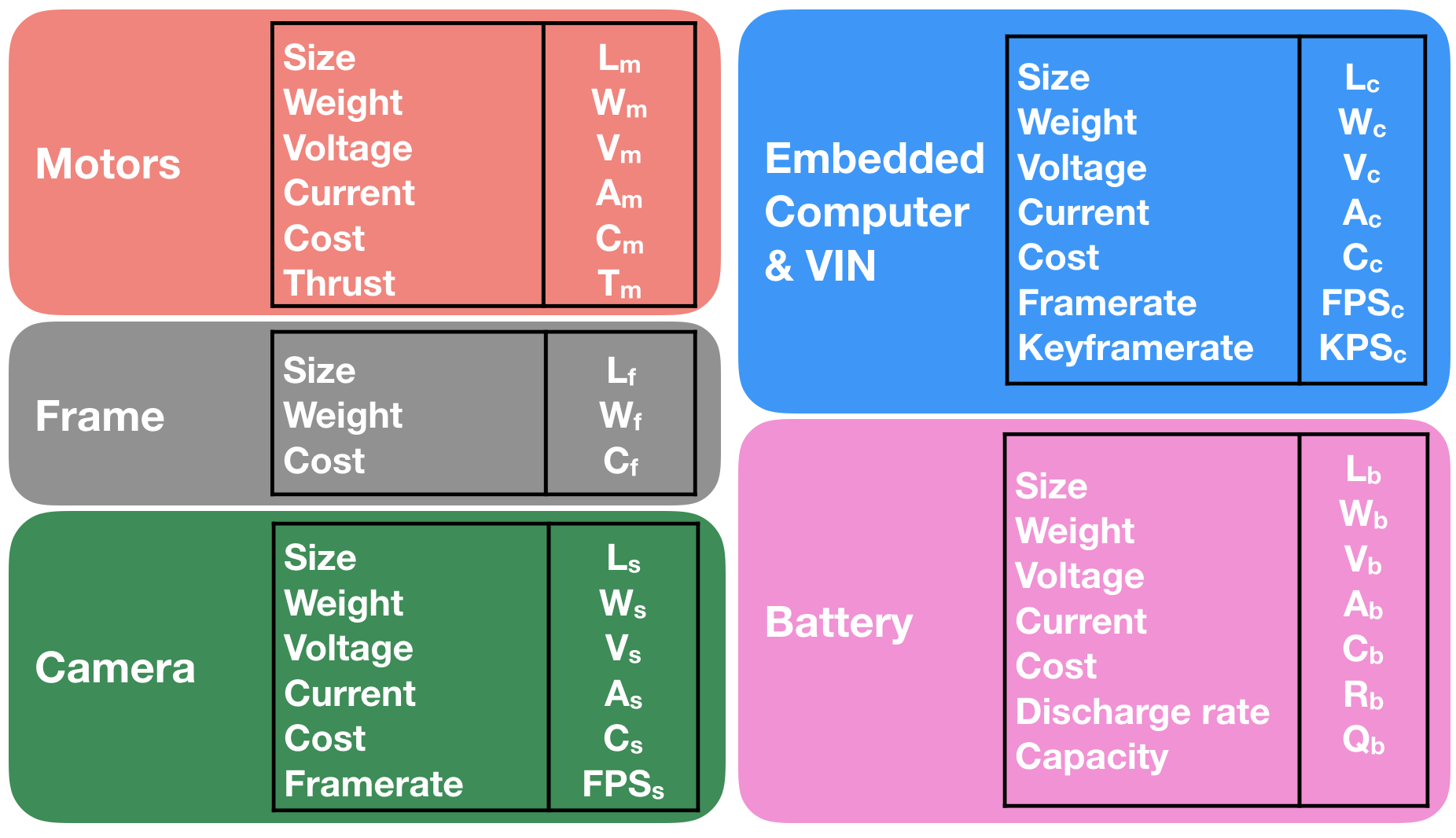}\vspace{-0.6cm}
    \caption{Drone co-design overview: modules and features. \label{fig:droneDesign} } 
\end{center}
\vspace{-1cm}
\end{figure}

\longversion{
 An overview of the modules we design and their corresponding features is given in Fig.~\ref{fig:droneDesign}.
% we decided to select among different ``computation \& software combos'' 
 % rather than designing the computational board and the software separately. 
 % This is motivated by the fact that we can use statistics reported in other papers to 
 % build up a feature table for a computation \& software combo. 
% While in principle we can choose independently the 
For the sake of simplicity, in this toy example we preferred not to design other components.
For instance, we disregarded other algorithms running on the board, e.g., for control, which are typically less computationally demanding.
We also neglected the presence/design of voltage adapters and connectors, 
while we assumed that each choice of motors comes with a suitable choice of ESC (Electronic Speed Control boards) 
and propellers.}{
	An overview of the modules we design and their corresponding features is given in Fig.~\ref{fig:droneDesign}.
}

\myparagraph{System-level performance}
%%%%%%%%%%%%%%%%%%%% NOTES %%%%%%%%%%%%%%%%%%%% %%%%%%%%%%%%%%%%%%%% 
% \bit
% % https://klsin.bpmsg.com/how-fast-can-a-quadcopter-fly/
% \item maximum speed: $v_{\max} = (1-(\frac{mg}{T})^2)^{\frac{1}{4}} \cdot \frac{T \sqrt{2}}{\sqrt{\rho c_D A m g}}$ with $T = nr motors \cdot T_i$
% \item size: = size of frame
% \eit
%%%%%%%%%%%%%%%%%%%% %%%%%%%%%%%%%%%%%%%% %%%%%%%%%%%%%%%%%%%% 
The second step of the design process is to quantitatively define the system-level performance metrics (what is the ``best drone''?). Since we consider an autonomous drone racing application, the best drone is one that can complete a given track 
as quickly as possible, hence a system that can navigate at high speed.
% and accelerations (these are particularly important to navigate in confined spaces). 
Therefore, in this example, the system performance metric is the top speed of the drone. 
We mainly consider \emph{forward} speed, but the presentation can be extended to maximize agility and accelerations.

%a vector of 2 elements, including the maximum speed and acceleration.
In order to derive an expression for the top (forward) speed, we observe Fig.~\ref{fig:drone}(b) 
and note that at its top speed, the forward acceleration is zero, hence the 
horizontal component $T_h$ of the thrust $T$ must compensate the drag force (i) $T_h = D$, and the 
vertical component $T_v$ of the thrust must compensate the force of gravity (ii) $T_v = F_g = Mg$, where 
$M$ is the overall mass of the drone and $g$ is the acceleration due to gravity. 
Let us focus on (i), and note that  (iii) $T^2 = T_h^2 + T_v^2$ and that the drag can be modeled as 
 (iv) $D = \frac{1}{2} \rho c_d A v^2$, where $\rho$ is the air density, 
$c_d$ is the drag coefficient (we take $\rho = 1.2 \text{kg}/\text{m}^3$ and $c_d = 1.3$), $A$ is the cross-sectional area, and $v$ is the 
forward speed of the drone. 
Substituting (ii), (iii), and (iv) back into (i), we obtain:
% 1.2; % Air density [kg/m3], 
% cd =  1.3
% F_{D} is the drag force,
% {\displaystyle \rho } \rho  is the density of the fluid,[11]
% {\displaystyle v} v is the speed of the object relative to the fluid,
% {\displaystyle A} A is the cross sectional area, and
% {\displaystyle C_{D}} C_{D} is the drag coefficient – a dimensionless number.
\beq
\label{eq:v1}
\sqrt{T^2 - (Mg)^2} = \frac{1}{2} \, \rho \, c_d \, A \, v_{\max}^2
\eeq 
Now we note that the cross-sectional area can be approximated as 
$A = \sin(\beta) \ell^2 = \frac{Mg}{T} L_\frame^2$, where $\beta$ is the pitch angle (basic trigonometry 
shows $\sin(\beta)  = \frac{Mg}{T}$), and $L_\frame$ is the length of the frame (approximated as a square).
Substituting this expression for $A$ in~\eqref{eq:v1} and rearranging the terms we obtain:
\beq
v_{\max} = \left( 
\frac{4 T^2}{ \rho^2 c_d^2 L_\frame^4}
 \left( \frac{T^2}{(Mg)^2} - 1 \right)
 \right)^{\frac{1}{4}}
\label{eq:v2}
\eeq 
Finally, assuming that we design a quadrotor, the cumulative thrust is the sum of the thrusts $T_\motor$ provided by each of the four motors, while the 
mass is the sum of the weight $W_i$ of all the modules $i \in \calM$: 
\beq
v_{\max} = \left( 
\frac{4 (4 T_\motor)^2}{ \rho^2 c_d^2 L_\frame^4}
 \left( \frac{(4 T_\motor)^2}{(g\sum_{i\in\calM} \omega_i W_i)^2} - 1 \right)
 \right)^{\frac{1}{4}}
\label{eq:vel}
\tag{SP}
\eeq 
% (we assume a quadrotor)
 where $\omega_i = 4$ if $i=\motor$ (we have 4 motors on a quadrotor, each one weighting $W_\motor$), or $\omega_i=1$ otherwise.

In order to make explicit that the values of $T_\motor, W_i, L_\frame$ depend on our design,  
we observe that $T_\motor = [\MF_\motor]_T \vxx_\motor$,  $W_i = [\MF_i]_W \vxx_i$
$L_\frame = [\MF_\frame]_L \vxx_\frame$, which makes~\eqref{eq:vel} a function of our design vector. 
Eq.~\eqref{eq:vel}  represents the system performance metric our design has to maximize. 
% We will discuss the modules we design below, while 
It is apparent from~\eqref{eq:vel} that the design 
% will influence the mass of the drone ($M$), its size ($\ell$), and the choice of motors will influence the 
% maximum thrust ($T$).
encourages drones which are small and lightweight ($L_\frame$ and $W_i$ appear at the denominator) 
and with large thrust ($T_\motor$  appears at the numerator).

%%%%%%%%%%%%%%%%%%%%%%%%%%%%%%%%%%%%%%%%%%%%%%%%%%%%%%%%%%%%%%%%%%%%%%%%%%%%%%%%%%%%%%%%%%%%%%%%%%%%%%%%%%%%%%%%%%%%%
\myparagraph{System-level constraints}
System constraints provide further specifications (in the form of hard constraints) on the task the drone is designed for.
For our drone example, we consider two main constraints: monetary budget and flight time.

\begin{enumerate}[wide, labelwidth=!, labelindent=0pt]
	\item \emph{Budget}: 
% The budget constraint is straightforward to model. 
Given a monetary budget $\bar{b}$, the budget constraint 
enforces that the sum of the costs $C_i$ of each module $i$ is within the budget:
\beq
\textstyle\sum_{i \in \calM} \omega_i C_i \leq \bar{b}
\label{eq:cost}
\tag{SC1}
\eeq
where again $\omega_i = 4$ if $i=\motor$ or $\omega_i=1$ otherwise.
% where we neglected manufacturing costs to keep things simple.

\item \emph{Flight time}: 
% The constraint on the flight time is also fairly straightforward to model. 
Given a minimum flight time $\bar{T}$ (this would be between 5-10 minutes
 in a real application), and calling $Q_\battery$ the battery capacity, and $A_i$ the average Ampere drawn by the $i$-th component (all assumed to 
 operate at the same voltage), then the time it takes to drain the battery must be $\geq \bar{T}$:
 \beq
 \frac{\alpha Q_\battery }{ \sum_{i \in \calM} \omega_i A_i } \geq \bar{T}
\label{eq:flightTime}
\tag{SC2}
 \eeq 
 where $\omega_i = 4$ if $i=\motor$ (again, we have 4 motors on a quadrotor), or $\omega_i=1$ otherwise;
 the constant $\alpha \in (0,1]$ (typically chosen to be around $0.8$) models the fact that we might not want to fully discharge 
our battery (e.g., LiPo batteries may be damaged when discharged below a recommended threshold).
\end{enumerate}

We remark that while flight time and budget constraints already make for an interesting problem, 
one can come up with many more system constraints, e.g., size and weight limits to 
participate into a specific drone racing competition\longversion{, or motor power constraints for safety 
or regulatory constraints.}{.}

\myparagraph{Implicit (module-level) constraints}
The implicit constraints make sure that each module can operate correctly and it is compatible with the other modules in the system.
We identified five implicit constraints:
\begin{enumerate}[wide, labelwidth=!, labelindent=0pt]
	\item \emph{Minimum thrust}: the cumulative thrust provided by the four motors has to be sufficient for flight. 
				Calling $T_\motor$ the thrust provided by each motor, and $W_i$ the weight of the $i$-th module, the minimum thrust constraint 
				can be written as:
				\beq
				4 T_\motor \geq \bar{r} \; g \sum_{i \in \calM} \omega_i W_i 
				\label{eq:IC1}
				\tag{IC1}
				\eeq
				where $\bar{r}$ is a given minimum thrust-weight ratio ($\bar{r}=2$ in this example), to ensure that 
				the thrust is sufficient to maneuver with agility, besides allowing the drone to hover. %taking off.
				% thrust over weight ratio for flight: thrust(motors) $\geq$ r( weight(motors) + weight(camera) + weight(computer) + weight(frame) + weight(battery) ) (r is the minimum ratio) 
	\item \emph{Power}: the battery should provide enough power to support the four motors, the camera, and the computer:
				\beq
				A_\battery V_\battery \geq 4 A_\motor V_\motor +  A_\camera V_\camera + A_\computer V_\computer
				\label{eq:IC2}
				\tag{IC2}
				\eeq
				where $A_i, V_i$ are the current and voltage for the $i$-th module.
				% power: power(battery) $\geq$ power(motors) + power(camera) + power(computer)
	\item \emph{Size}: all components should fit on the frame. Assuming we can stack components on top of each other, 
			the size constraint enforces that the size of each component should be smaller than the size of the frame: 
			\beq
			L_\frame \geq L_i \quad \text{with } i \in \{\motor,\camera,\computer,\battery\}
			\label{eq:IC3}
			\tag{IC3}
			\eeq 
			% In principle, we might also: constraints: size to pass through obstacles
			% size: size(frame) $\geq$ size(battery) + size(computer) + size(camera)
	\item \emph{Minimum camera frame-rate}: the framerate of the camera should be fast enough to allow tracking visual features, during fast 
			motion. This is necessary for the visual-inertial navigation (VIN) algorithms to properly estimate position and attitude of the drone. Assuming that our VIN 
			front-end can track a feature moving by at most $\delta_u$ pixels between frames, and corresponding to the projection of a 
			3D point at distance $d$ from the camera (we assume $\delta_u = 30$ pixels, $d = 5$ meters), the frame-rate $\FPS_s$ of the camera is bounded by~\cite{Zhang17rss-vioChip}:
			\beq
			\FPS_s \geq \frac{f v_{\max}}{ \delta_u  d }
			\label{eq:IC4}
			\tag{IC4}
			\eeq
			where $f$ is the focal length of the camera, and $v_{\max}$ is the maximum speed of the drone, given in~\eqref{eq:vel}. 
			% frame-rate lower bound (From VIO chip paper): $\tau_{f} \leq \frac{ \delta_u  d }{f v_{\max}}$ where $d = 1$m
			 % is the minimum distance from an object to the camera, $f$ is the focal length and $\delta_u$ is the maximum pixel displacement that 
			 % can be tracked by our feature tracker. 
	\item \emph{Minimum VIN frame-rate}: the VIN algorithm operating on the 
			embedded computer on the drone should be able to process each frame, hence the VIN frame-rate (recall that we select from a catalog of VIN+computer combinations) must be larger than the %th is lower bounded by the 
			camera frame-rate:
			\beq
			\FPS_c \geq \FPS_s
			\label{eq:IC5}
			\tag{IC5}
			\eeq
			%  \item framerate of vio frontend should be larger than framerate of camera to keep up: $\omega_{VIO-frontend} > \omega_c$
	% \item \emph{minimum VIN keyframe-rate}: we consider VIN algorithms based on keyframes. For those, we need to impose that most 
	% 		features can be tracked (i.e., they do not disappear from the image) between keyframes. This imposes another constraint 
	% 		on how frequently we have to process a keyframe, which in turn has implication on how fast our computer should be.
	% 		Calling $\KPS_c$ the keyframe rate for a computer $c$, we obtain the following lower bound (same derivation as in~\cite{Zhang17rss-vioChip}): 
	% 		\beq
	% 		\KPS_c \geq \frac{f v_{\max}}{ \Delta_u  d }
	% 		\label{eq:IC6}
	% 		\tag{IC6}
	% 		\eeq
	% 		where $\Delta_u$ is the maximum pixel displacement 
	%  \item keyframe-rate lower bound (From VIO chip paper): $\tau_{kf} \leq \frac{ \Delta_u  d }{f v_{\max}}$ where $d = 1$m
 % is the minimum distance from an object to the camera, $f$ is the focal length and $\Delta_u$ is the maximum pixel displacement that 
 % such that most pixels remain in the image between keyframes (e.g., $\Delta_u = 1/3$ image size).
\end{enumerate}
% 
% we neglected the battery discharge rate: current drawn by each component should be less than the discharge rate of the battery
%
% Approximations:
% - only consider power drawn by motors
% - only consider weight of batteries for drag
%
% Quadcopter flight times =(Battery Capacity * Battery Discharge /Average Amp Draw)*60
%
% Battery Capacity: For calculator you have to take the battery’s capacity in amp hours, To convert from mAh to Ah, battery capacity divide by 1000. For instance, 1800mAh/1000=1.8Ah.
%
% Battery Discharge: It’s common practice to not discharge your LiPo batteries below 20% mAh during flight; In other words, the effective capacity is only 80% that can be used during flight time. For instance, 1.8Ah*0.8=1.44A 
%
% Average Amp Draw: Before quadcopter battery calculator you work out the average amp draw, you have to know two things, one is about your carrying weight of your quadcopter which include battery weight. An other thing is the parameters of quadcopter motor, Read your motor instructions, you have to know how many amps one motor will be draw to produce 100g of thrust? This is the key points in the calculation of Average Amp Draw. 

\myparagraph{Implementation and results} 
% Dual variables tell that VIO is the constraint on speed
% \myparagraph{Results: trade-offs}
% plot performance for different budget choices
The ``optimal'' drone design solves the following optimization problem:
\beal
\;\;\;\max_{\vxx \in \calX} & \text{(SP)}\\
\subject & \text{(SC1)}, \text{(SC2)} \\
& \text{(IC1)}, \text{(IC2)}, \text{(IC3)}, \text{(IC4)}, \text{(IC5)}
\label{eq:droneCodesign}
\eeal
Most of the constraints are already in a form that is amenable for our approach
(Sec.~\ref{sec:algorithms}).  Only the objective~\eqref{eq:vel} and the
constraints~\eqref{eq:flightTime} and~\eqref{eq:IC4} have a more involved expression, which we
further develop in \longversion{the appendix,}{the supplemental material~\cite{},} where we show how
to approximate those expressions in a form that fit a linear co-design solver.

\newcommand{\mpw}{4cm}
\begin{figure}[t]
\begin{minipage}{\textwidth}
\begin{tabular}{cc}%
\begin{minipage}{\mpw}%
\centering
\includegraphics[width=1.06\columnwidth]{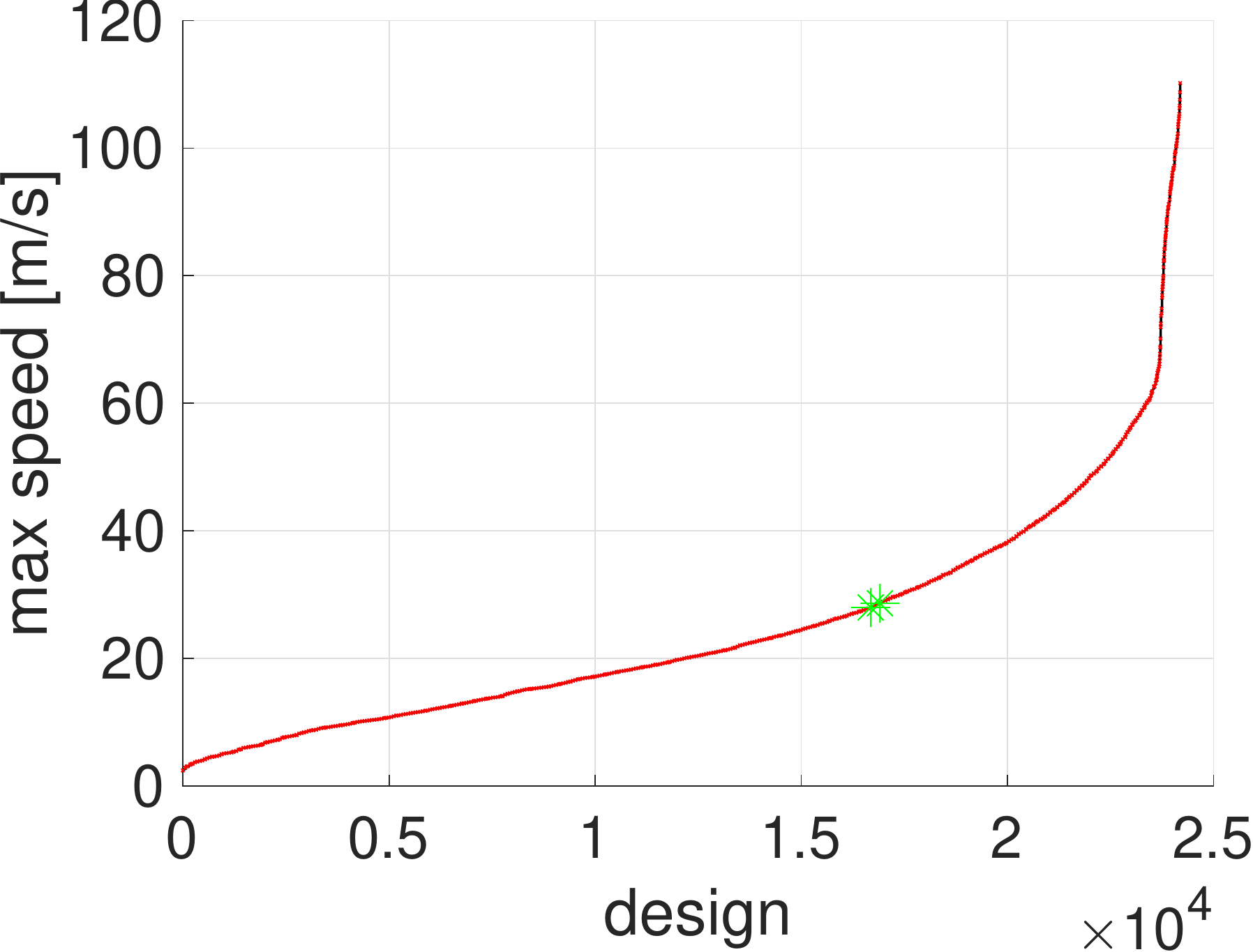} \\
(a)
\end{minipage}
& 
\begin{minipage}{\mpw}%
\centering%
\includegraphics[width=1.06\columnwidth]{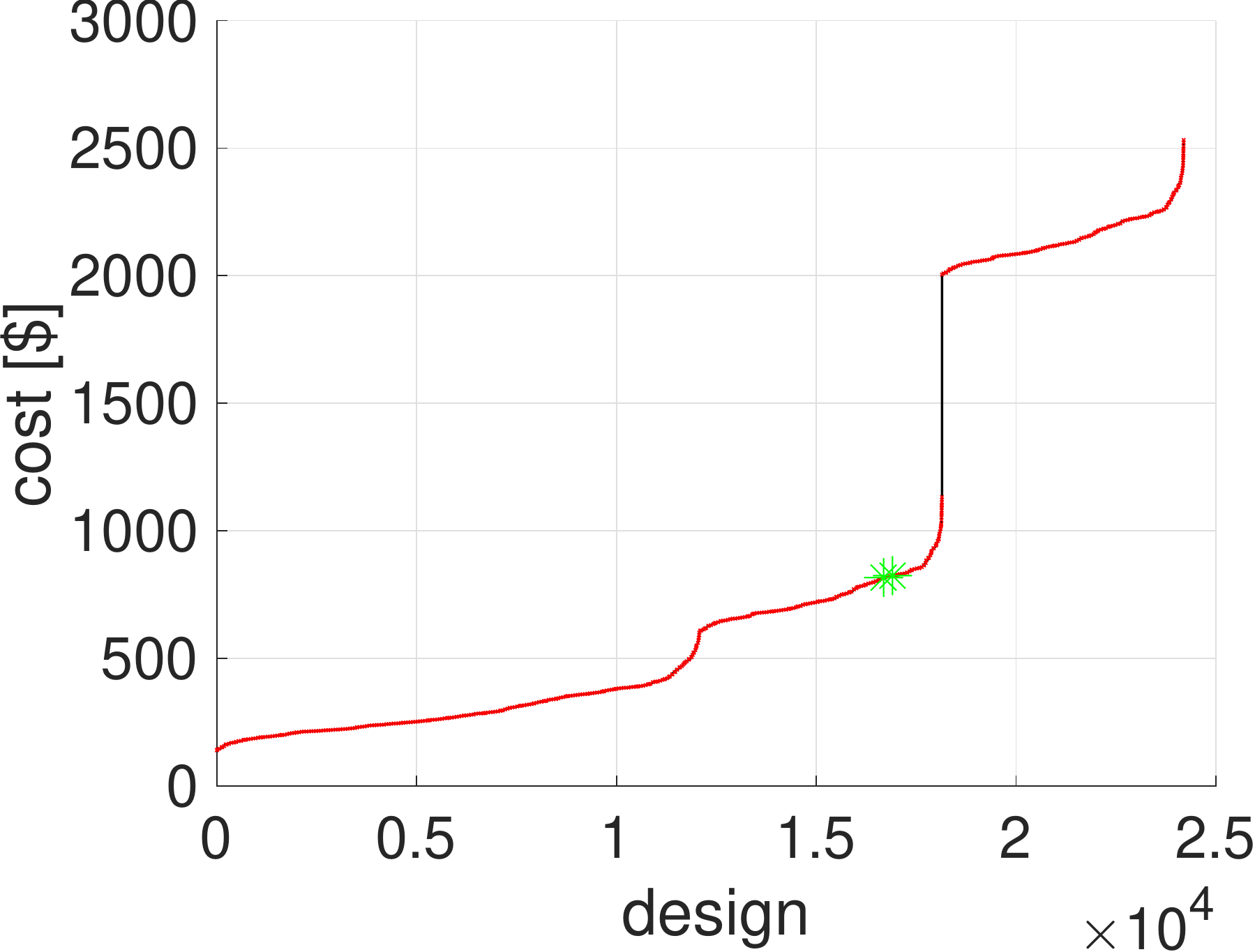} \\
(b)
\end{minipage}
\end{tabular}
\end{minipage}%
\caption{\label{fig:combinations}
(a) Estimated maximum speed and (b) cost for any potential configuration of modules in our catalog. 
Infeasible designs are marked in red, while feasible designs are marked with a green star.
\vspace{-1cm}
}
\end{figure}

The optimal design we found in our implementation suggests that an optimal configuration of modules would include 
a Stormer 220 FPV Racing Quadcopter Frame Kit, EMAX1045 motors, an NVIDIA TX2 computer, a 60 frame-per-second camera, and  
 a Tattu 5100mAh 3S 10C Battery Pack. The cost of such a drone would be $\$700.84$, and the drone would have a flight 
 time of $10.42$ minutes and a top speed of $35.52$ meters per second, which is compatible with the performance 
 one expects from a racing drone~\cite{drone17}. 
 CPLEX was able to find an optimal design in $0.3$ seconds.
 Since the design space is relatively small we can compute the estimated maximum speed and the cost of every potential combination 
 of the modules in our catalog. These results are shown in Fig.~\ref{fig:combinations}, where we also report whether the configuration is 
feasible (it satisfied all system and implicit constraints) or not. The figure shows that there are indeed only four feasible designs 
in our catalog all attaining fairly similar performance and cost.
% * max velocity [m/s]: 35.5259
% * max flight time [minutes]: 10.4235
% * cost [dollars]: 700.841
%We implemented the following optimization pro
% repositories of components already exist
% \url{https://www.ecalc.ch/}
%
% use state of the art model in Propeller Thrust and Drag in Forward Flight
% Rajan Gill and Raffaello D’Andrea
%
% Towards A Swarm of Agile Micro Quadrotors
% Alex Kushleyev, Daniel Mellinger, Vijay Kumar GRASP Lab, University of Pennsylvania
% constraints: size to pass through obstacles, 
% performance: agility, speed
% speed vs tracking capabilities?

%%% Local Variables:
%%% mode: latex
%%% TeX-master: "../main"
%%% End:
}{\input{experiments-drone_shorter}}

\newcommand{\teamsize}{R}
\newcommand{\push}{P}
\newcommand{\weight}{W}
\newcommand{\robotset}{\{1,\ldots,K\}}

% ==========================================================
\subsection{Codesign of Heterogeneous Multi-Robot Teams}
\label{sec:experiments-multiRobot}

\begin{figure}[t]
  \centering % ,natwidth=610,natheight=642
  \includegraphics[width=.22\textwidth]{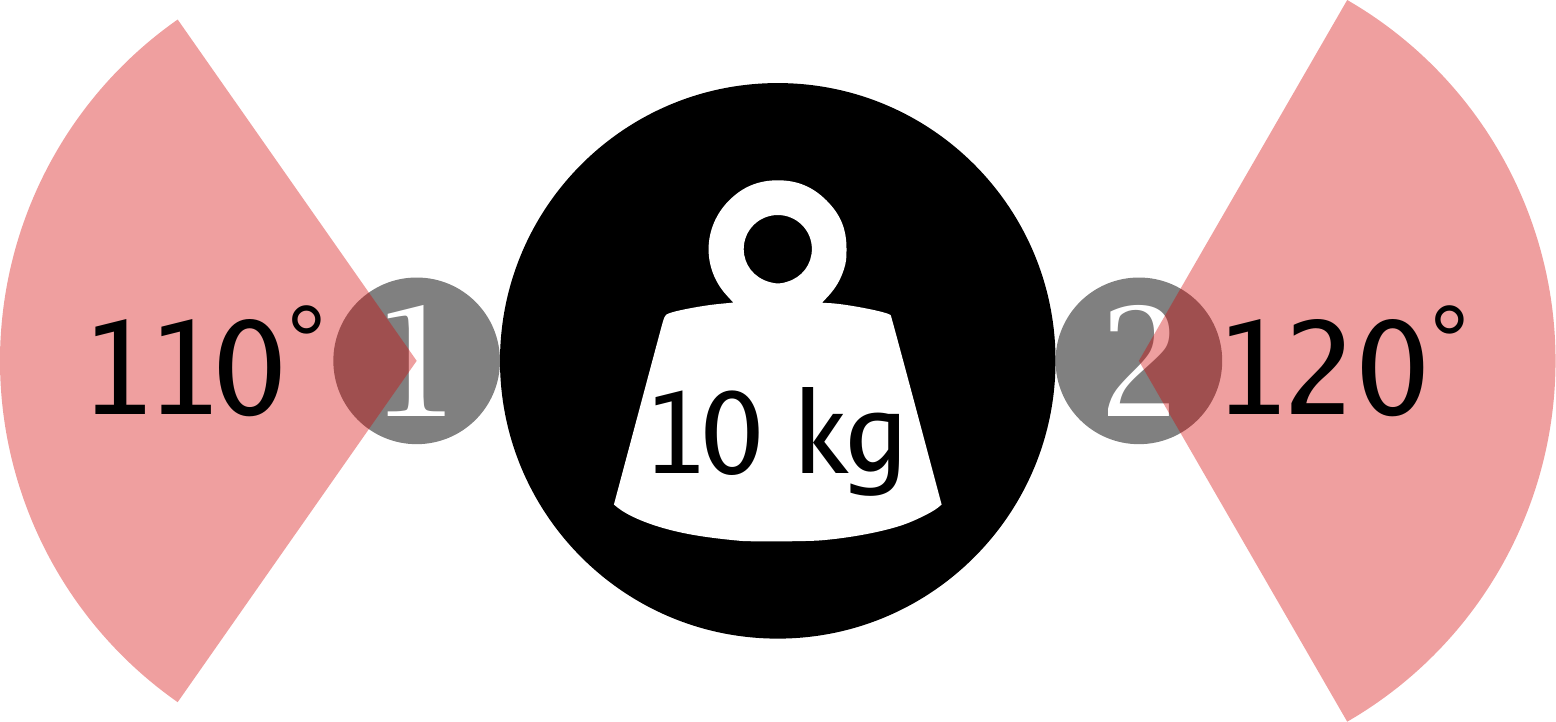}
  \includegraphics[width=.22\textwidth]{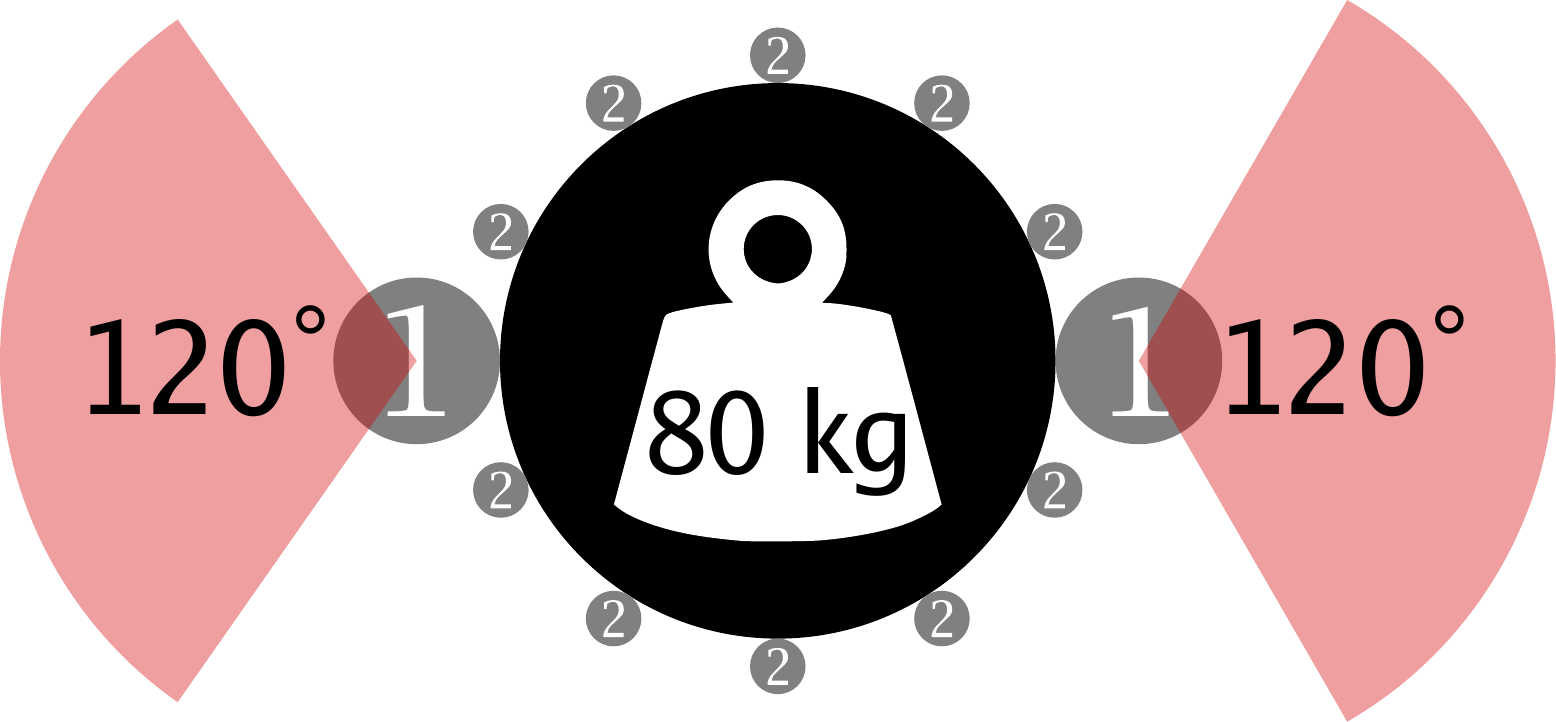}
  \vspace{-5mm}
  \caption{Collective transport example (top view), where multiple robots (gray circles) carry a large object (black circle).
  The figure shows two solutions found by CPLEX when the object weighs $\unit[10]{kg}$ (left) and $\unit[80]{kg}$ (right).
  % Two solutions found with the approach discussed in
  %   Sec.~\ref{sec:experiments-multiRobot}. The left solution is for a multi-robot system that
  %   carries a \unit[10]{kg} object; second if for an \unit[80]{kg} object. In the left solution, the
  %   robots differ in their sensor coverage, and are otherwise identical. In the right solution, the
  %   robots specialize in two types: type 1 is larger and performs carrying and sensing; type 2 is
  %   smaller and only performs carrying. In addition, type 2 robots are lighter and offer a better
  %   push margin than type 1 robots.
    }
    \vspace{-5mm}
  \label{fig:multirobotSolutions}
\end{figure}

This section applies the proposed framework to the co-design of a heterogeneous multi-robot team.
In particular, we consider a \emph{collective transport scenario}, in which the robots must
collectively carry a heavy object while avoiding obstacles. The robots must be configured to allow
for efficient carrying and for wide sensor coverage, while battery power and frame size constrain
the capabilities that any single robot possesses. The end result of our design is a heterogeneous
team, composed of robots that specialize in carrying and robots that specialize in sensing. As a
simplifying assumption, we focus on objects and robots with circular shapes (see
Fig.~\ref{fig:multirobotSolutions}).

\myparagraph{Modules} We consider four types of modules: \emph{frames} (\frame), \emph{sensors} (\sensor),
\emph{motors} (\motor), and \emph{batteries} (\battery). Frames' features include size and weight; sensors' features include
% characterized by 
coverage (as a percentage of the surrounding area), size, weight, and power
consumption; motors' features include weight, size, power consumption and force exerted; and
batteries' features include size, weight, and power generated. 
% \longversion{Note that we consider a slightly
% higher level of abstraction for the modules with respect to the drone example.}{Note that we consider a slightly
% higher level of abstraction for the modules with respect to the drone example.}

Since we have to design multiple robots, we use the notation $\vxx^k_{i}$ to denote the design
vector associated to the $i$-th module of robot $k$.  As before, we use $[\vxx^k_{i}]_j$ to denote the $j$-th entry of $\vxx^k_{i}$.
 $[\vxx^k_{i}]_j = 1$ if we chose the $j$-th element in the catalog $\calC_i$ of module $i$ on robot $k$, or zero otherwise. 
% , where $j$ is the
% index of a specific feature choice (i.e., an entry of $\vxx^k_{i}$). 
We let $k \in \robotset$ where 
%and, in
% this scenario, 
we calculate the upper bound $K$ by considering the maximum number of robots that can
encircle the object when the smallest frame is used. If the radius of the object is
$R^{\text{object}}$ and the radius of the smallest frame is $R^{\text{min frame}}$, then
\begin{equation}
\label{eq:maxTeamSize}
 K = \left\lfloor\frac{\pi (R^{\text{object}} + R^{\text{min frame}})}{R^{\text{min frame}}}\right\rfloor.
\end{equation}

\myparagraph{System-level performance} In a multi-robot system, performance is inherently
non-monotonic. In Hamann's analysis~\cite{hamann_towards_2012}, performance is expressed as the
ratio of two components: cooperation $C(\vxx)$ and interference $I(\vxx)$. Cooperation refers to
those phenomena that contribute to the task at hand; interference corresponds to the phenomena that
diminish the system performance. A simple approach to capture both aspects is to cast the co-design
problem as an \emph{interference minimization} problem, while using cooperation measures as system
constraints. In~\cite{hamann_towards_2012}, $I(\vxx)$ is expressed as an exponential that decays
with the size of the team $\teamsize$, from which it follows %. Taking the logarithm: 
% and recalling that maximizing a function is the same as minimizing its negation:
%and recalling that ``$\max s$'' is the same as ``$\min -s$'' we can write

{%\small
\begin{equation}
\label{eq:spmulti}
\min_{\vxx \in \calX} I(\vxx) \equiv \min_{\vxx \in \calX} \e^{\teamsize(\vxx)} \equiv \min_{\vxx \in \calX} \teamsize(\vxx) \equiv \max_{\vxx \in \calX} -\teamsize(\vxx)
  % \fnP(\vxx) = - \teamsize(\vxx)
\end{equation}}

\noindent
where the symbol ``$\equiv$'' denotes that these changes of objective do not alter the solution of the optimization problem.
 In~\eqref{eq:spmulti}
 we also observed that the team size is a function of our design ($\teamsize = \teamsize(\vxx)$), i.e., the design algorithm can decide to use 
less robots than the upper bound~\eqref{eq:maxTeamSize}. Eq.~\eqref{eq:spmulti} shows that our design will attempt to 
use the least number of robots, in order to minimize interference.
To capture $\teamsize(\vxx)$, we augment $\vxx$ with a binary element (later called the ``slot'') that indicates
whether the features of a robot $k$ are being used or not (in other words: if the robot is part of the team of not). We can thus define
$\teamsize(\vxx) = \sum_{k=1}^K \vxx^k_{\text{slot}}$ and write
\begin{equation}
  \fnP(\vxx) = -\textstyle\sum_{k=1}^K \vxx^k_{\text{slot}}.
  \label{eq:mrperftransport}
  \tag{SP}
\end{equation}

\myparagraph{System-level constraints} We consider two families of system constraints concerning motors
and sensing.
\begin{enumerate}[wide, labelwidth=!, labelindent=0pt]
\item \emph{Push for object transportation:} 
the team as a whole must be able to carry the
  object; this is essentially the cooperation measure mentioned above.  
  % To model this constraints, 
  For this, we need that the push $\push_j$ exerted by the choice of motor $j$ exceeds the sum of the weights $\weight_{i,j}$ of 
  each choice modules ($i \in \calM$ and $j \in\calC_i$) plus the weight $W$ of the object to transport:
  \begin{equation}
    \sum_{ k \in \robotset,j \in \calC_\motor} \hspace{-0.6cm} \push_j [\vxx^k_{\motor}]_j \geq  \hspace{-5mm}
    \sum_{k \in \robotset,i \in \MM, j \in \calC_i}  \hspace{-1cm} \weight_{i,j} [\vxx^k_{i}]_j + W
    \label{eq:mrconstrSC1}
    \tag{SC1}
  \end{equation}
  In addition, each robot's motor push must exceed the weight of the robot itself:
  \begin{equation}
    \sum_{j \in \calC_\motor} \push_j [\vxx^k_{\motor}]_j  \geq \!\!\!
    \sum_{i \in \MM, j \in \calC_i} \!\!\! \weight_{i,j} [\vxx^k_{i}]_j \quad \forall k \in \robotset 
    \label{eq:mrconstrSC2}
    \tag{SC2}
  \end{equation}
\item \emph{Sensor coverage:} the robot sensors must ensure that at least 50\% of the area around
  the object is covered at any time:
  \begin{equation}
    \longversion{}{\textstyle} \sum_{j \in \calC_\sensor} S_j [\vxx^k_{\sensor}]_j \geq 50\%
    \label{eq:mrconstrSC3}
    \tag{SC3}
  \end{equation}
  where $S_j$ is the coverage provided by the $j$-th sensor choice. % of sensors.
\end{enumerate}

\begin{table}[t]
  \centering
  \caption{Implicit constraints for the multi-robot co-design example.}
  \label{tab:multirobotImplConstraints}
  \begin{tabular}{m{3.2cm}m{3.2cm}r}
    \textbf{Meaning} & \textbf{Formalization} & \textbf{Tag} \\
    \hline
    \hline
    If a slot is activated, the robot has one frame & $\forall k \quad \vxx^k_{\text{slot}} = \sum_{j \in \calC_\frame} [\vxx^k_{\frame}]_j$ & (IC1)\\
    \hline
    If a slot is activated, the robot has one motor & $\forall k \quad \vxx^k_{\text{slot}} = \sum_{j \in \calC_\motor} [\vxx^k_{\motor}]_j$ & (IC2)\\
    \hline
    If a slot is activated, the robot has one battery & $\forall k \quad \vxx^k_{\text{slot}} = \sum_{j \in \calC_\battery} [\vxx^k_{\battery}]_j$ & (IC3)\\
    \hline
    If a slot is activated, the robot has at most one sensor & $\forall k \quad \vxx^k_{\text{slot}} \geq \sum_{j \in \calC_\sensor} [\vxx^k_{\sensor}]_j$ & (IC4)\\
    \hline
    The power consumption of a robot is lower than the power given by the battery & $\forall k \quad \sum_{j \in \calC_\battery} P_{\battery,j} [\vxx^k_{\battery}]_j \geq \sum_{i \in \{\motor,\sensor\},j \in \calC_i} P_{i,j} [\vxx^k_{i}]_j$ & (IC5)\\
    \hline
    The total size of the components of a robot is lower than the size of the chassis & $\forall k \quad \sum_{j \in \calC_\frame} A_{\frame,j} [\vxx^k_{\frame}]_j \geq \sum_{i \in \{\motor,\sensor,\battery\},j \in \calC_i} A_{i,j} [\vxx^k_{i}]_j$ & (IC6)\\
    \hline
    \hline
  \end{tabular}
  \vspace{-5mm}
\end{table}

\myparagraph{Implicit (module-level) constraints} %The constraints reported in
Table~\ref{tab:multirobotImplConstraints} summarizes the implicit constraints in our example.
% must be imposed to ensure the correctness of the
% formulation. 
The symbol $P_{i,j}$ denotes the power offered by battery $j$ or used by the $j$-th choice of module $i$.
% $(i,j)$; 
$A_{i,j}$ denotes the area offered by a frame $j$ or the area used by the $j$-th choice of module $i$.
% $$used by a component $(i,j)$.

\myparagraph{Implementation and results} To form the catalog of possible modules, we considered 10
alternatives for each type of module and 2 alternatives for the size of the robot frame, for a total
of $2,000$ combinations. Using the size and weight of the object to transport, we explored the space
of optimal solutions, which involved teams of up to $K=65$
robots. Fig.~\ref{fig:multirobotSolutions} reports the solutions we found by solving two instances
of the problem, where the object to carry weighted $\unit[10]{kg}$ and $\unit[80]{kg}$,
respectively.  In the left diagram, CPLEX concluded that two large robots are sufficient to carry a
$\unit[10]{kg}$ object. The robots are equipped with identical motors and batteries, and differ only
in their sensor coverage. In the right diagram, CPLEX generated a solution including two types of
robots: the larger type offers a lower pushing margin, but is capable of wide sensing; the smaller
type is lighter and offers a higher pushing margin, but it performs no sensing. The presence of
robots that carry no sensors allows them to collectively shoulder the bulk of the pushing. The
interested reader can find the complete CPLEX implementation at
\url{https://bitbucket.org/lucacarlone/codesigncode/}.

%%% Local Variables:
%%% mode: latex
%%% TeX-master: "../main"
%%% End:

\section{Conclusion}
\label{sec:conclusion}
We presented an approach for \emph{computational robot co-design} that formulates the joint
selection of the modules composing a robotic system in terms of mathematical programming.  While our
approach is rooted in the general context of binary optimization, we discussed a number of
properties---specific to our co-design problem---that allow rephrasing several co-design problems
including non-linear functions of the features of each module in terms of \emph{binary linear
  programming} (\BLP).  Modern \BLP algorithms and implementations can solve problems with few
thousands of variables in seconds, which in turn allows attacking interesting co-design problems.
We demonstrated the proposed co-design approach in two applications: the design of an autonomous
drone and the design of a multi-robot team for collective transportation. Future work includes
extending the set of functions for which we can solve the co-design problem in practice, and adding
continuous variables (e.g., wing length, 3D-printed frame size) as part of the co-design problem.

% \begin{itemize}
% \item time varying specifications
% \item connessione con Linear temporal logic
% \end{itemize}

%%% Local Variables:
%%% mode: latex
%%% TeX-master: "../main"
%%% End:

\longversion{
\appendix

\subsection{Drone Co-design}
\label{sec:appendix-drone}

This section discusses how to approximate the objective~\eqref{eq:vel} and the constraints~\eqref{eq:flightTime} and~\eqref{eq:IC4}
using binary linear functions. The following paragraphs 
deal with each case.

\myparagraph{A linear lower bound for the objective~\eqref{eq:vel}}
The design has to maximize the maximum forward speed, which, from~\eqref{eq:vel}, has the following expression:
\beq
v_{\max} = \left( 
\frac{4 (4 T_\motor)^2}{ \rho^2 c_d^2 L_\frame^4}
 \left( \frac{(4 T_\motor)^2}{(g\sum_{i\in\calM} \omega_i W_i)^2} - 1 \right)
 \right)^{\frac{1}{4}}
\label{eq:vel_a1}
\eeq 
This expression does not fall in cases (a), (b), (c) in Section~\ref{sec:algorithms}; 
moreover, it involves all modules, hence taking the approach (d) of Section~\ref{sec:algorithms} would be impractical 
(it would simply lead to enumerating every possible design choice, implying a combinatorial explosion of the state space).

The approach we take in this section is to approximate Problem~\eqref{eq:droneCodesign} by replacing its objective with a 
\emph{linear lower bound}. We remark that, as shown in Section~\ref{sec:algorithms}, while the design is required to be 
linear in $\vxx$, it can be heavily nonlinear in the features $\MF_i$. 
In the following we show how to obtain a linear lower bound  for~\eqref{eq:vel_a1}. 
For this purpose, we observe that from~\eqref{eq:IC1}, the thrust-weight ratio must be larger than $\bar{r}$, 
hence $(4 T_\motor) / (g\sum_{i\in\calM} \omega_i W_i) \geq \bar{r}$. Therefore, it holds: 
\bea
v_{\max} = \left( 
\frac{4 (4 T_\motor)^2}{ \rho^2 c_d^2 L_\frame^4}
 \left( \frac{(4 T_\motor)^2}{(g\sum_{i\in\calM} \omega_i W_i)^2} - 1 \right) \right)^{\frac{1}{4}} \nonumber  \\ 
 \geq
\left( 
\frac{4 (4 T_\motor)^2}{ \rho^2 c_d^2 L_\frame^4}
 \left( \bar{r}^2- 1 \right) \right)^{\frac{1}{4}} 
  = \kappa \frac{T_\motor^\frac{1}{2}}{L_\frame} = \kappa \frac{([\MF_\motor]_T \, \vxx_\motor)^\frac{1}{2}}{[\MF_\frame]_L \vxx_\frame} %\nonumber
 \label{eq:vel_a2}
\eea 
where $\kappa$ is a constant, irrelevant for the maximization. 
The function~\eqref{eq:vel_a2} now falls in the case (c) discussed in Section~\ref{sec:algorithms} and 
can be made linear in $\vxx$ by taking the logarithm.
% \bea
% v^4 = \frac{T^2}{c A^2} \left(\left(\frac{T}{mg} \right)^2 - 1\right) \geq \frac{T^2}{c A^2} (r-1)
% \eea
% where $r$ is the minimum thrust over weight ratio. 
% Therefore, maximizing $v$ is the same as maximizing its log (and dropping constants):
% \bea
% \check{v} = 2\log(T) - 2\log(A)
% \eea
% assume focal length same for all cameras

\myparagraph{A conservative linear approximation for~\eqref{eq:IC4}}
% \myParagraph{Upper Bound for Frame-rate}
Similarly to the case discussed above, we approximate the ``problematic'' constraints via surrogates that are linear in $\vxx$. 

The approach we take is to substitute a constraint in the form $f(\vxx) \leq 0$ with a linear constraint $\hat{f}(\vxx) \leq 0$, 
where $f(\vxx) \leq \hat{f}(\vxx)$ for any $\vxx \in \calX$. This guarantees that for any $\vxx$ that makes $\hat{f}(\vxx) \leq 0$, then 
also $f(\vxx) \leq 0$, i.e., we still guarantee to compute feasible (but potentially more conservative) designs.

Therefore, in the following we show how to obtain a linear \emph{upper} bound  for~\eqref{eq:IC4}. 
Let us start by writing~\eqref{eq:IC4} more explicitly, by substituting $v_{\max}$ from~\eqref{eq:sc}:  
\beq
\FPS_\camera \geq \frac{f}{ \delta_u  d } \left( 
\frac{4 (4 T_\motor)^2}{ \rho^2 c_d^2 L_\frame^4}
 \left( \frac{(4 T_\motor)^2}{(g\sum_{i\in\calM} \omega_i W_i)^2} - 1 \right)
 \right)^{\frac{1}{4}}
\label{eq:IC4_app1}
\eeq
or equivalently (taking the $4$-th power of each member):
\beq
\FPS_\camera^4 \geq \frac{4 f^4}{ \delta_u^4  d^4 \rho^2 c_d^2 } 
\frac{(4 T_\motor)^2}{ L_\frame^4}
 \left( \frac{(4 T_\motor)^2}{(g\sum_{i\in\calM} \omega_i W_i)^2} - 1 \right)
\label{eq:IC4_app1}
\eeq
Taking the logarithm of both sides we obtain:
\bea
4 \log(\FPS_\camera) \geq \beta + 2\log(4 T_\motor) - 4 \log(L_\frame) + \nonumber \\
+ \log\left( \frac{(4 T_\motor)^2}{(g\sum_{i\in\calM} \omega_i W_i)^2} - 1 \right)
\eea
where we defined the constant $\beta \doteq \log( \frac{4 f^4}{ \delta_u^4  d^4 \rho^2 c_d^2 })$.

Rearranging the terms:
\bea
2\log(4 T_\motor) - 4 \log(L_\frame) + \nonumber \\
+ \log\left( \frac{(4 T_\motor)^2}{(g\sum_{i\in\calM} \omega_i W_i)^2} - 1 \right) - 4 \log(\FPS_\camera) +\beta \leq 0
\label{eq:IC4_app2}
\eea
We note that so far we did not take any approximation since each operation we applied ($4$-th power, logarithm, reordering) 
preserves the original inequality. Moreover,~\eqref{eq:IC4_app2} is in the form $f(\vxx) \leq 0$. 
In the rest of this paragraph, we show how to compute a linearized upper bound for $f(\vxx)$.
% Therefore, to produce a conservative approximation of the constraint~\eqref{eq:IC4_app2} we need to find a 
% linear upper bound for the function in the left-hand-side. 
For this purpose we note that the following 
chain of inequalities holds:
\bea
2\log(4 T_\motor) - 4 \log(L_\frame) +  \nonumber\\
+ \log\left( \frac{(4 T_\motor)^2}{(g\sum_{i\in\calM} \omega_i W_i)^2} - 1 \right) - 4 \log(\FPS_\camera) +\beta \!\!&\!\!\!\! \overset{(i)}{\leq} \!\!\!\! \nonumber
\\
2\log(4 T_\motor) - 4 \log(L_\frame) + \nonumber\\
+ \log\left( \frac{(4 T_\motor)^2}{(g\sum_{i\in\calM} \omega_i W_i)^2} \right) - 4 \log(\FPS_\camera) +\beta \!\!&\!\!\!\! \overset{(ii)}{=} \!\!\!\! \nonumber
\\
4\log(4 T_\motor) - 4 \log(L_\frame) + \nonumber\\
- 2\log\left( \sum_{i\in\calM} g \,\omega_i\, W_i \right) - 4 \log(\FPS_\camera) +\beta \!\!&\!\!\!\! \overset{(iii)}{\leq} \!\!\!\! \nonumber
\\
4\log(4 T_\motor) - 4 \log(L_\frame) + \nonumber\\
- 2 \sum_{i\in\calM} \frac{1}{|\calM|} \log\left( g \,\omega_i\, W_i \right) \!-\! 2\log(|\calM|) \!-\! 4 \log(\FPS_\camera) +\beta
\label{eq:IC4_app3}
\eea
where in (i) we dropped the $-1$ and used the fact that the logarithm is a non-decreasing function, 
in (ii) we simply developed the logarithm of the ratio, 
and in (iii) we used the Jensen's inequality:
\bea
\log\left(\frac{\sum_{i=1}^n y_i}{n} \right) \geq  \frac{\sum_{i=1}^n \log(y_i)}{n} \iff \\
\log\left( \sum_{i=1}^n y_i \right) \geq  \sum_{i=1}^n \frac{1}{n}\log(y_i) + \log(n)
\eea
which holds for any integer $n$ and positive scalar $y_i$.

Making the dependence of $T_\motor$, $L_\frame$, $W_i$, and $\FPS_\camera$ on the design vector $\vxx$ explicit in~\eqref{eq:IC4_app3}, we obtain:
\bea
\hat{f}(\vxx) \doteq 4\log(4 [\MF_\motor]_T \vxx_\motor) - 4 \log([\MF_\frame]_L \vxx_\frame) + \nonumber\\
- 2 \sum_{i\in\calM} \frac{1}{|\calM|} \log\left( g\,\omega_i\, [\MF_i]_W \vxx_i \right)  + \nonumber\\
- 2\log(|\calM|) - 4 \log([\MF_\camera]_\FPS \vxx_\camera) +\beta
\label{eq:IC4_app4}
\eea
which now falls in the case (b) discussed in Section~\ref{sec:algorithms} and 
can be easily made linear with respect to $\vxx$.
\begin{comment}
\bea
v^4 = \frac{T^2}{c A^2} \left(\left(\frac{T}{mg} \right)^2 - 1\right) \leq  \frac{T^4}{c A^2 (mg)^2} \leq \frac{T^4}{c A^2 (m_b g)^2} = \hat{v}
\eea
where $m_b$ is the mass of the battery (recall that $m = \sum_i m_i$ where $m_i$ is the mass of each module).

Now for the framerate we need $\tau_{f} \leq \frac{ \delta_u  d }{f v_{\max}}$ or equivalently $\omega_{f} \geq \frac{ f }{\delta_u  d } v_{\max}$ 
where $\omega$ is the framerate. A necessary condition for this eq to be satified is:
\beq
\omega_{f}^4 \geq (\frac{ f }{\delta_u  d } \hat{v})^4 = k^4 \frac{T^4}{c A^2 (m_b g)^2}
\eeq
where we defined $k \doteq \frac{ f }{\delta_u  d }$.
Taking the log at both sides:
\beq
4\log(\omega_{f}) \geq 4\log(k) + 4 \log(T) - \log(c) - 2\log(A) - 2\log(m_b g)
\eeq
recall that $T = 4 T_m$ where $T_m$ is the thrust produced by each motor:
\beq
4\log(\omega_{f}) \geq 4\log(k) + 4 \log(4 T_m) - \log(c) - 2\log(A) - 2\log(m_b g)
\eeq
\end{comment}

\myparagraph{A conservative linear approximation for~\eqref{eq:flightTime}}
Similarly to the previous section, here we approximate the ``problematic'' constraint~\eqref{eq:flightTime} via surrogates that are linear in $\vxx$. 
In particular, as before, we substitute a constraint in the form $f(\vxx) \leq 0$ with a linear constraint $\hat{f}(\vxx) \leq 0$, 
where $f(\vxx) \leq \hat{f}(\vxx)$ for any $\vxx \in \calX$. This guarantees that for any $\vxx$ that makes $\hat{f}(\vxx) \leq 0$, then 
also $f(\vxx) \leq 0$, i.e., we still guarantee to compute feasible (but potentially more conservative) designs.

Let us start by reporting the original expression we want to approximate~\eqref{eq:flightTime}:  
\beq
 \frac{\alpha Q_\battery }{ \sum_{i \in \calM} \omega_i A_i } \geq \bar{T}
\label{eq:flightTime_app1}
 \eeq 
which, again, does not fall in cases (a), (b), (c) in Section~\ref{sec:algorithms}, and 
involves all modules, hence making the approach described in case (d) of Section~\ref{sec:algorithms} impractical.

In order to obtain a conservative approximation of the constraint~\eqref{eq:flightTime_app1}, we first take the logarithm
of both members:
\beq
 \log(\alpha Q_\battery) - \log\left( \sum_{i \in \calM} \omega_i A_i \right) \geq \log(\bar{T})
\label{eq:flightTime_app2}
 \eeq 
 which does not alter the inequality since all involved quantities are positive.
 Rearranging the terms:
\beq
\log\left( \sum_{i \in \calM} \omega_i A_i \right) - \log(\alpha Q_\battery) + \log(\bar{T}) \leq 0
\label{eq:flightTime_app3}
 \eeq 
 We can find an upper bound for the left-hand-side of~\eqref{eq:flightTime_app3} as follows:
\bea
\log\left( \sum_{i \in \calM} \omega_i A_i \right) - \log(\alpha Q_\battery) + \log(\bar{T}) \leq \nonumber \\
\log( 6 A_\motor) - \log(\alpha Q_\battery) + \log(\bar{T})
\label{eq:flightTime_app4}
 \eea 
where we used $\sum_{i \in \calM} \omega_i A_i = A_\camera + A_\computer + 4 A_\motor \leq 6 A_\motor$ (note: frame and batteries 
are assumed to draw zero current)
%and the fact that $\max_i(A_i) = A_\motor$, i.e., 
which follows from the fact the motors are the modules that draw more current.
% In our current implementation we take an even more direct approach and approximate: 
Therefore we can substitute the original constraint~\eqref{eq:flightTime_app1} with a conservative linear approximation:
\bea
\hat{f}(\vxx) \doteq 
\log( 6  [\MF_\motor]_A \vxx_\motor) - \log(\alpha [\MF_\battery]_Q \vxx_\battery) + \log(\bar{T}) \leq 0
\label{eq:flightTime_app4}
 \eea 
which now falls in the case (b) discussed in Section~\ref{sec:algorithms} and 
can be easily made linear with respect to $\vxx$.

\begin{comment}
Equilibrium of forces along the horizontal (forward) direction produces:
\bea
T_h - D = ma
\eea
where $T_h$ is the horizontal component of the thrust and $D$ denotes the rotor drag.
Developing the expression above we obtain:
\bea
\sqrt{T^2 - (mg)^2} - \frac{1}{2} \rho c_d A_h v^2  = ma
\eea
where $A_h$ is the effective area for the drag computation and $A_h = A \frac{mg}{T}$ where $A$ is the area of the quadrotor:
At maximum speed we have no acceleration:
\bea
\frac{1}{2} \rho c_d A \frac{mg}{T} v^2 = \sqrt{T^2 - (mg)^2} 
\eea
squaring both terms and defining $c \doteq (\frac{1}{2} \rho c_d)^2$
\bea
c A^2 (\frac{mg}{T})^2 v^4 = T^2 - (mg)^2 
\eea
leaving only $v$ on the lhs:
\bea
v^4 = \frac{T^2}{c A^2} \left(\left(\frac{T}{mg} \right)^2 - 1\right)
\eea
\end{comment}

}{}

% Bibliography
\bibliographystyle{IEEEtran}
\bibliography{refs,myRefs,zotero}

\end{document}